%% file: main.tex
\documentclass[sigplan,10pt,nonacm,screen]{autotailor-acmart}
\makeatletter
\renewcommand\acmBadgeR[2][]{%
  \ifx\@acmBadgeR\@empty\else
    \g@addto@macro{\@acmBadgeR}{\hspace{\@ACM@badge@skip}}%
  \fi
  \g@addto@macro{\@acmBadgeR}{%
    \smash{\raisebox{-8pt}{%
      \href{#1}{\includegraphics[width=\@ACM@badge@width]{#2}}}}}}
\makeatother
\usepackage{subcaption}
\usepackage{listings}
\usepackage{lstfiracode}
\usepackage{enumitem}
\usepackage{soul}

\ActivateVerbatimLigatures

\PassOptionsToPackage{prologue,dvipsnames}{xcolor}
\PassOptionsToPackage{bookmarksnumbered,unicode,colorlinks}{hyperref}
\usepackage{amsmath}
\usepackage{pifont}
\usepackage{multirow}
\usepackage{algorithm}
\usepackage{algpseudocode}
\usepackage[dvipsnames]{xcolor}
\usepackage{caption}
\usepackage{subcaption}
\usepackage{graphbox}
\usepackage{placeins}
\lstdefinelanguage{TOML}{
  morekeywords={[2][arch],[var.global_vars],[var.stage_vars],[var.block_vars]},
  sensitive=true,
  morestring=[b]",
  morecomment=[l]\#,
}

\lstdefinestyle{toml}{
  language=TOML,
  basicstyle=\ttfamily\footnotesize,
  keywordstyle=\color{blue}\bfseries,
  keywordstyle={[2]\color{purple}\bfseries},
  stringstyle=\color{green!50!black},
  numberstyle=\tiny\color{gray},
  commentstyle=\color{gray},
  moredelim=[l][\color{blue}]{=},
  literate=
    *{0}{{{\color{orange}0}}}{1}
     {1}{{{\color{orange}1}}}{1}
     {2}{{{\color{orange}2}}}{1}
     {3}{{{\color{orange}3}}}{1}
     {4}{{{\color{orange}4}}}{1}
     {5}{{{\color{orange}5}}}{1}
     {6}{{{\color{orange}6}}}{1}
     {7}{{{\color{orange}7}}}{1}
     {8}{{{\color{orange}8}}}{1}
     {9}{{{\color{orange}9}}}{1}
     {.0}{{{\color{orange}.0}}}{2},
 frame=single,
  rulecolor=\color{gray!70}, 
  frameround=tttt,
  aboveskip=2pt,
  belowskip=2pt
}
\usepackage[colorlinks]{hyperref}

\algtext*{EndWhile}
\algtext*{EndIf}
\algtext*{EndFor}
\algtext*{EndFunction}

\fancypagestyle{firstpagestyle}{\fancyhf{}}
\microtypesetup{protrusion=false}

\acmBadgeR{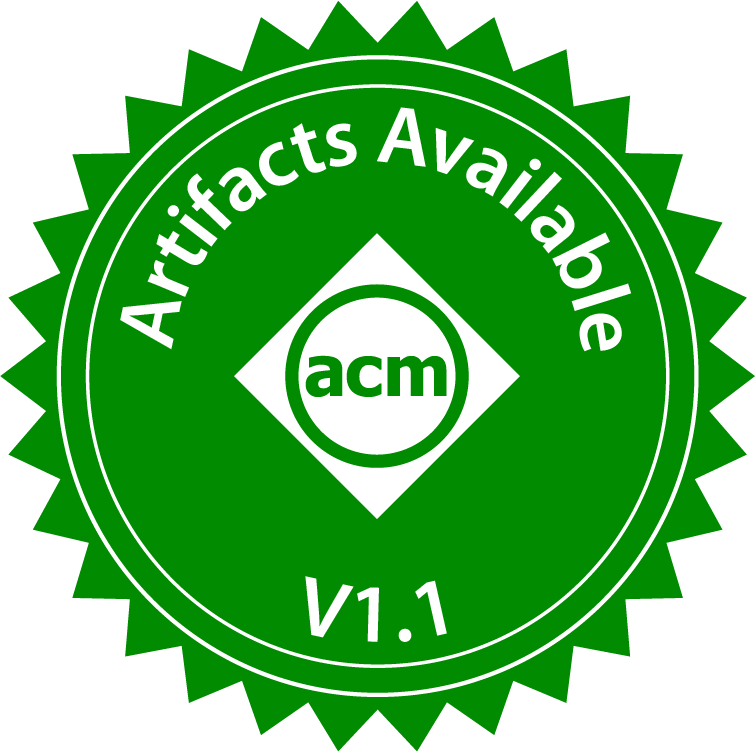}
\acmBadgeR{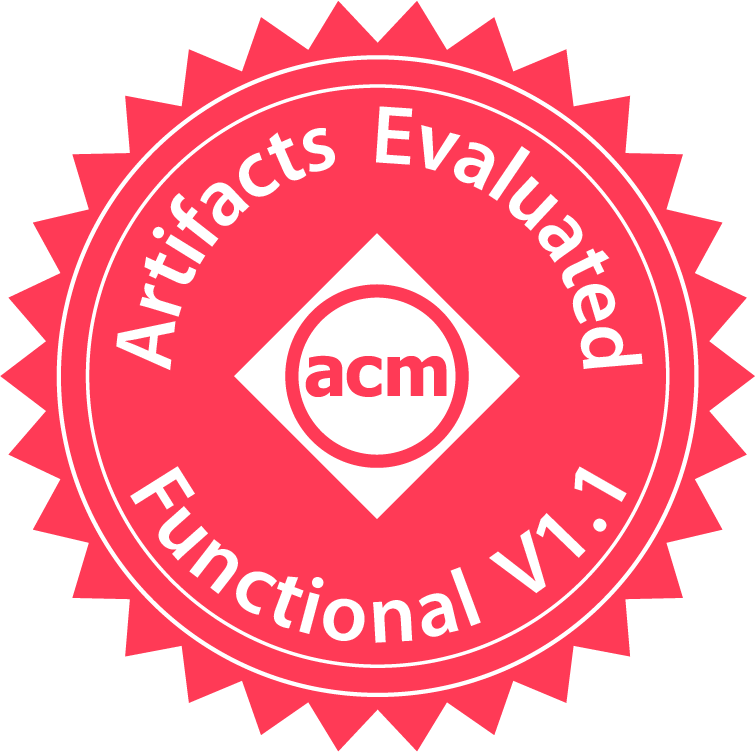}
\acmBadgeR{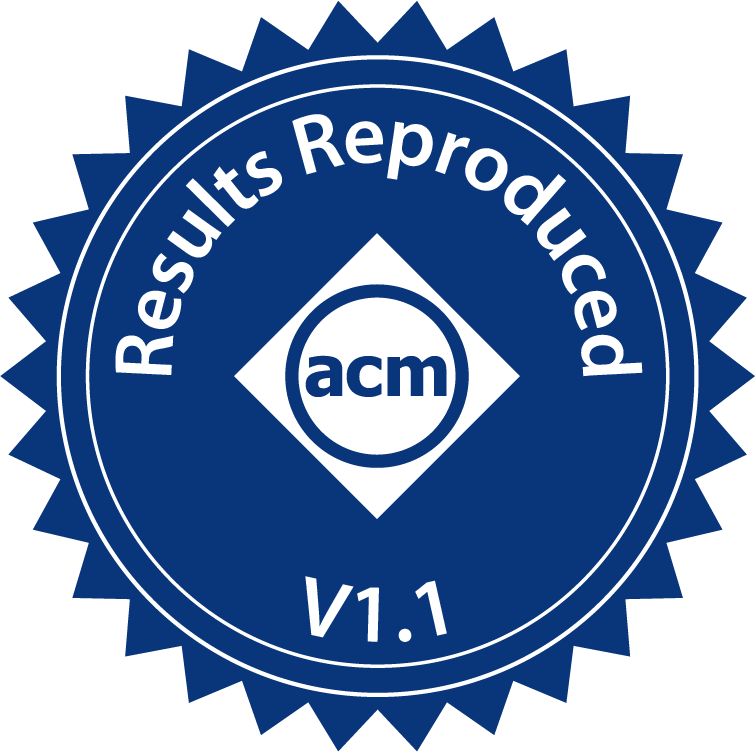}

\begin{document}

\title{AutoTailor: Automatic and Efficient Adaptive Model Deployment for Diverse Edge Devices}

\settopmatter{authorsperrow=5}
\author{Mengyang Liu}
\affiliation{\institution{Southeast University}\city{}\country{}}
\affiliation{\institution{University of Macau}\city{}\country{}}

\author{Chenyu Lu}
\affiliation{\institution{HKUST (GZ)}\city{}\country{}}

\author{Haodong Tian}
\affiliation{\institution{Tsinghua University}\city{}\country{}}

\author{Fang Dong}
\authornote{Corresponding author: \href{mailto:fdong@seu.edu.cn}{fdong@seu.edu.cn}.}
\affiliation{\institution{Southeast University}\city{}\country{}}

\author{Ruiting Zhou}
\affiliation{\institution{Southeast University}\city{}\country{}}

\author{Wei Wang}
\affiliation{\institution{HKUST}\city{}\country{}}

\author{Dian Shen}
\affiliation{\institution{Southeast University}\city{}\country{}}

\author{Guangtong Li}
\affiliation{\institution{Southeast University}\city{}\country{}}

\author{Ye Wan}
\affiliation{\institution{Southeast University}\city{}\country{}}

\author{Li Li}
\affiliation{\institution{University of Macau}\city{}\country{}}

\input{1_abstract.tex}

\maketitle
\pagestyle{empty}

\input{2_intro.tex}

\input{3_back.tex}

\input{4_overview.tex}

\input{5_design.tex}

\input{7_impl}

\input{8_evaluation.tex}

\FloatBarrier
\input{9_relatedwork.tex}
\input{10_conclusion.tex}

\bibliographystyle{ACM-Reference-Format}
\bibliography{ref}

\end{document}

%% file: 1_abstract.tex
\begin{abstract}

On-device machine learning (ML) has become a fundamental component of emerging mobile applications.
Adaptive model deployment delivers efficient inference for heterogeneous device capabilities and performance requirements through customizing neural architectures.
SuperNet-based approaches offer a promising solution by generating numerous model variants from a pre-trained ML model.
However, applying SuperNet in existing frameworks requires tedious model-aware development and time-consuming hardware-aware profiling, which limits their practical adoption.

We present AutoTailor, the first framework to enable automated, end-to-end SuperNet-based adaptive model deployment for edge devices.
Unlike manual SuperNet construction, AutoTailor employs a computation graph-guided compilation approach to automatically transform user-provided ML models into SuperNets.
To support efficient specialization, AutoTailor incorporates learning-free latency and accuracy predictors, enabling low-cost yet accurate performance prediction.
Our extensive evaluations demonstrate that AutoTailor reduces the lines of code for SuperNet construction by 10.8--34.5$\times$, reduces predictor profiling time by 2.2--11.5$\times$, and achieves up to 15.60\% absolute accuracy improvement and 60.03\% latency reduction compared to state-of-the-art approaches across diverse models and devices.
AutoTailor's source code is publicly available at \url{https://github.com/liu-mengyang/AutoTailor}.

\end{abstract}

%% file: 2_intro.tex
\section{Introduction}

Deep neural networks (DNNs) are increasingly executed on edge devices to provide private intelligence for a wide range of mobile applications, including augmented reality \cite{aria}, video analytics \cite{remix}, and robot navigation \cite{panopticus}.
Static DNN deployment often fails to meet the diverse requirements of applications (e.g., real-time responsiveness or energy efficiency) on heterogeneous edge devices (e.g., industrial GPU boards or personal mobile phones).
To solve this problem, adaptive model deployment \cite{adaptivenet,legodnn,nestdnn} has emerged as a widely adopted strategy.
It first generates multiple model variants with different latency-accuracy trade-offs from a pre-trained DNN and then deploys the most suitable one according to device capabilities and runtime conditions.

\begin{figure}[tb]
    \centering
    \includegraphics[width=\columnwidth]{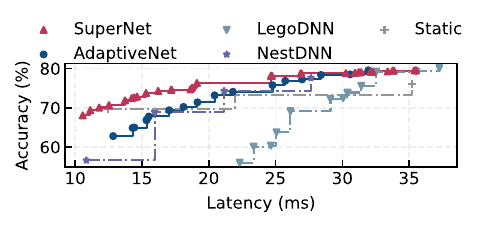}
    \vspace{-30pt}
    \caption{Latency-accuracy tradeoff of adaptive ResNet-50 on the Exynos 1380 SoC's big cores using the ImageNet1k dataset across different methods. SuperNet consistently achieves state-of-the-art performance. Static uses ResNet-18 and ResNet-34 as smaller model variants. }
    \label{fig:figure1}
    \vspace{-10pt}
\end{figure}
The number and granularity of model variants determine the performance of adaptive model deployment.
Early work adopted coarse-grained uniform filter pruning \cite{nestdnn} to generate variants with different widths.
LegoDNN \cite{legodnn} refined this approach by applying non-uniform pruning, providing more variants with improved overall performance.
AdaptiveNet \cite{adaptivenet} further extended the design space by varying not only width but also depth.
These methods generate model variants through \textit{block scaling}, where each block has multiple weight-independent variants, and complete model variants are constructed by combining them.
However, in block scaling approaches, each block requires fine-tuning to preserve accuracy, which limits scalability in producing a larger number of variants and achieving higher performance.

To overcome these limitations, \textit{SuperNets} \cite{ofa,elasticvit,hat} offer a scalable alternative through weight sharing.
Instead of training model variants independently, SuperNets jointly train multiple variants (SubNets) that share parameters.
As shown in Fig.~\ref{fig:figure1}, adopting SuperNets for adaptive deployment achieves up to $34.67\%$ latency reduction with the same accuracy and $7.89\%$ absolute accuracy improvement under the same latency over state-of-the-art adaptive deployment methods for the ResNet50 DNN \cite{resnet}.

Unfortunately, applying SuperNets in practice incurs substantial development and deployment costs due to their high flexibility.
On the development side, users must manually transform a static DNN into a dynamic one that supports multiple architectural modifications. This process not only bloats the codebase but also demands deep expertise in DNN architecture design.
For example, ElasticViT \cite{elasticvit} transforms LeViT \cite{levit} with 4$\times$ more lines of code than its original implementations (1638 vs. 406) to make each module transformable.
Retiarii~\cite{retiarii} alleviates this burden through the \textsc{Mutator} abstraction, eliminating the need to manually implement transformable operators.
Users can annotate target operators with predefined \textsc{Mutator}s and specify candidate configurations, enabling automatic model variant generation.
Nevertheless, this process still requires architectural expertise to identify suitable operators and ensure configuration compatibility, particularly to prevent tensor shape mismatches.
Thus, SuperNet construction remains challenging and labor-intensive, especially for non-experts.

On the deployment side, optimizing model architectures for specific devices requires profiling candidate variants, which is highly time-consuming.
The vast number of variants within a SuperNet makes exhaustive offline profiling infeasible, while online profiling during optimization introduces significant inefficiency.
To mitigate this, existing approaches \cite{nnmeter,litepred} train operator-level latency predictors and aggregate their outputs to estimate model-level latency, thereby simplifying predictor construction.
However, the accuracy of these learning-based methods fundamentally depends on the quantity and quality of training data.
Consequently, building such predictors commonly demands hours of data collection (e.g., 1 hour of profiling to build a predictor with only 50\% accuracy as shown in Fig.~\ref{fig:figure7}) for each model-platform pair, yet the resulting predictors still suffer from non-negligible prediction errors.

Our goal is to enable automatic and efficient SuperNet deployment, thereby advancing adaptive model deployment in practice.
Achieving this goal presents two key challenges.
First, automating SuperNet construction requires decoupling users from the DNN programs, leaving the framework unaware of which modules are transformable or what candidate choices to assign.
Second, the vast number of possible model variants makes accurate predictor training difficult: the larger the design space, the more profiling data are required to train the predictor effectively, resulting in prohibitive time costs.

We observe two fundamental characteristics of SuperNets that can be leveraged to address these challenges.
First, SuperNets exhibit \textbf{graph structural consistency}: they generate variants by applying architectural modifications to a base architecture, which preserves the computation graph of the original DNN.
This property enables transformable modules to be automatically identified and replaced directly from the computation graph, eliminating the need for manual annotation.
Second, SuperNets exhibit \textbf{cross-SubNet architectural similarity}: although they contain many SubNets, most operators are repeated.
Our analysis (Section~\ref{sec2.4}) shows a high repetition ratio (96--98\%), meaning the number of unique operators is significantly small.
This property enables learning-free predictors that profile only unique operators, eliminating costly sampling and predictor training.

We present AutoTailor, an end-to-end adaptive model deployment framework for edge devices.
AutoTailor automatically compiles supported static-graph DNNs into SuperNets represented by TailorIR, a novel intermediate representation.
The resulting SuperNet is then fine-tuned to preserve the accuracy of its model variants.
To enable efficient latency prediction, AutoTailor identifies unique operators across the SuperNet by eliminating redundant configurations through modification dependency analysis and execution-free tensor shape inference.
Operator-level latency measurements are then aggregated to predict model-level latency.
For accuracy estimation, AutoTailor applies a similar approach to improve prediction efficiency.
Finally, guided by these predictors and user-specified requirements, AutoTailor searches for the optimal model variant and generates deployable DNNs tailored to the target edge platform.

The key contributions of AutoTailor are as follows:
\begin{itemize}
    \item To the best of our knowledge, AutoTailor is the first framework to provide automated, end-to-end adaptive model deployment for edge devices.
    \item We propose a computation graph-guided compilation method and a method for building learning-free predictors to reduce development and deployment overheads in applying SuperNets.
    \item We extensively evaluate AutoTailor across diverse DNNs and edge devices, demonstrating that it reduces engineering effort by 10.8--34.5$\times$ and reduces predictor profiling time by 2.2--11.5$\times$, while achieving up to $15.60\%$ absolute accuracy improvement and $60.03\%$ latency reduction compared to state-of-the-art approaches.
\end{itemize}

%% file: 3_back.tex
\section{Background and Motivation}

\begin{figure}[t]
    \centering
    \includegraphics[width=\columnwidth]{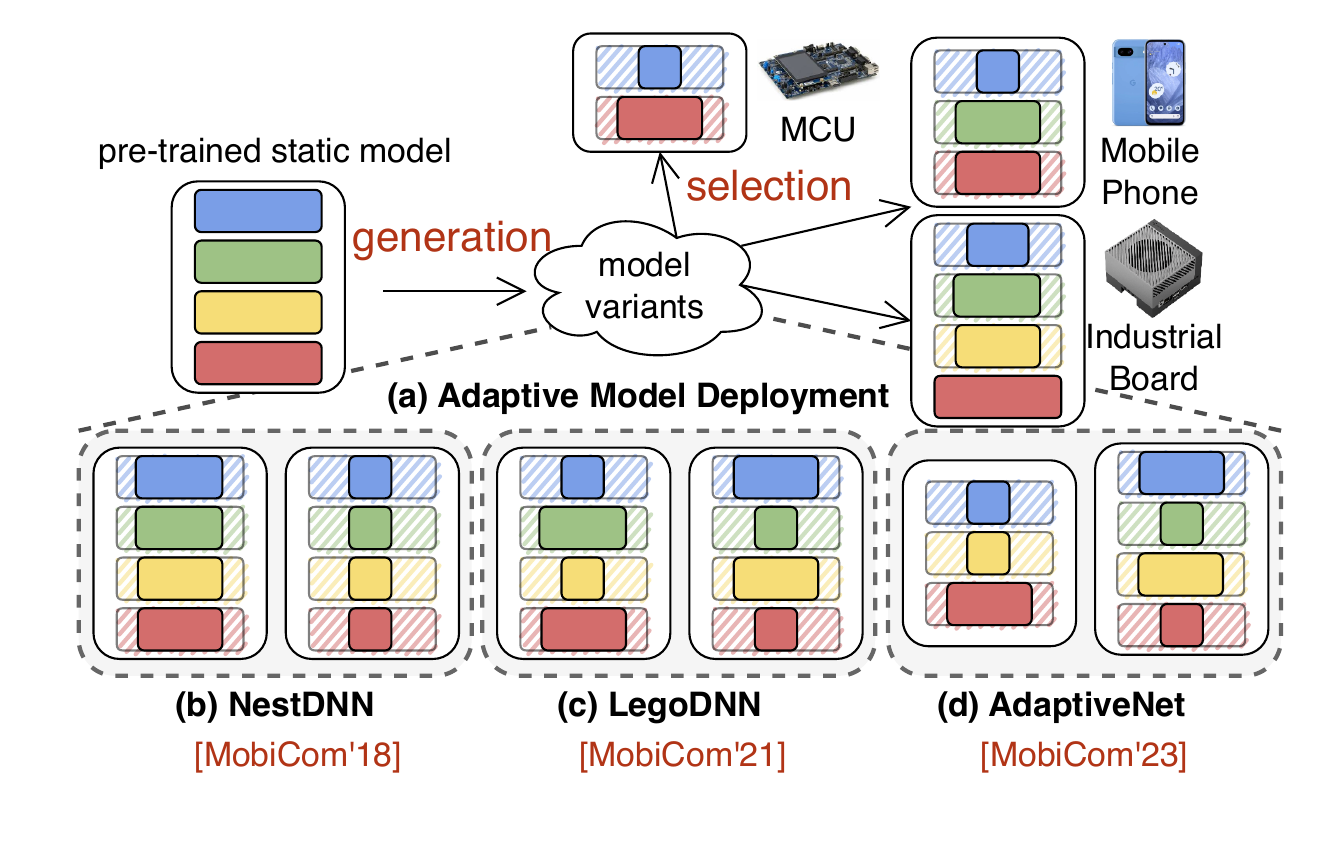}
    \vspace{-35pt}
    \caption{An illustration of adaptive model deployment and different model variant generation methods. A four-layer DNN is used as a running example, where each color denotes a distinct layer.}
    \label{fig:figure2}
    \vspace{-10pt}
\end{figure}

\subsection{Adaptive Model Deployment for Edge Devices}
\label{sec2.1}
Deploying powerful deep neural networks (DNNs) on edge devices to enable private and personalized intelligence has become a rapidly growing trend in mobile applications \cite{edgeai2025}.
Traditional static model deployment, placing a single fixed-architecture model on the device for inference, proves inadequate for edge platforms due to two key characteristics of edge devices.
First, edge devices exhibit wide hardware diversity, ranging from high-performance NVIDIA Orin boards \cite{nvidiaorin} to commodity smartphones \cite{snapdragonsoc} and resource-constrained microcontrollers (MCUs) \cite{stm32}.
Second, performance requirements vary across edge applications: some real-time processing tasks demand high-frequency DNN inference \cite{remix,elf,blastnet,aria}, while persistent background applications \cite{ember,intelbeyond} require energy-efficient, low-power DNN inference.
To better align model execution with heterogeneous resources (e.g., compute capacity) and application requirements (e.g., accuracy and latency), adaptive model deployment has emerged as a promising approach \cite{nestdnn,legodnn,adaptivenet}.
This process consists of two steps as shown in Fig. \ref{fig:figure2}(a): (1) generating multiple model variants with different architectures, accuracy, and efficiency; and (2) selecting the most suitable variant for a given device and runtime requirement.

The effectiveness of adaptive model deployment depends on the diversity and granularity of available model variants.
Fig.~\ref{fig:figure2}(b-d) illustrates representative variants generated by existing adaptive deployment methods.
Early work such as NestDNN~\cite{nestdnn} relied on uniform filter pruning~\cite{filterpruning}, applying the same sparsity ratio across all layers.
LegoDNN~\cite{legodnn} extended this idea by enabling non-uniform layer-wise pruning, thereby expanding the design space with greater flexibility.
Moving beyond width-only modifications,AdaptiveNet~\cite{adaptivenet} further introduced a layer-skipping strategy, which substantially enlarges the search space and refines the granularity of architectural adaptation.
Collectively, these efforts reveal a consistent trend toward expanding the model variant design space to improve adaptive deployment performance.

\begin{figure}[t]
    \centering
    \includegraphics[width=\columnwidth]{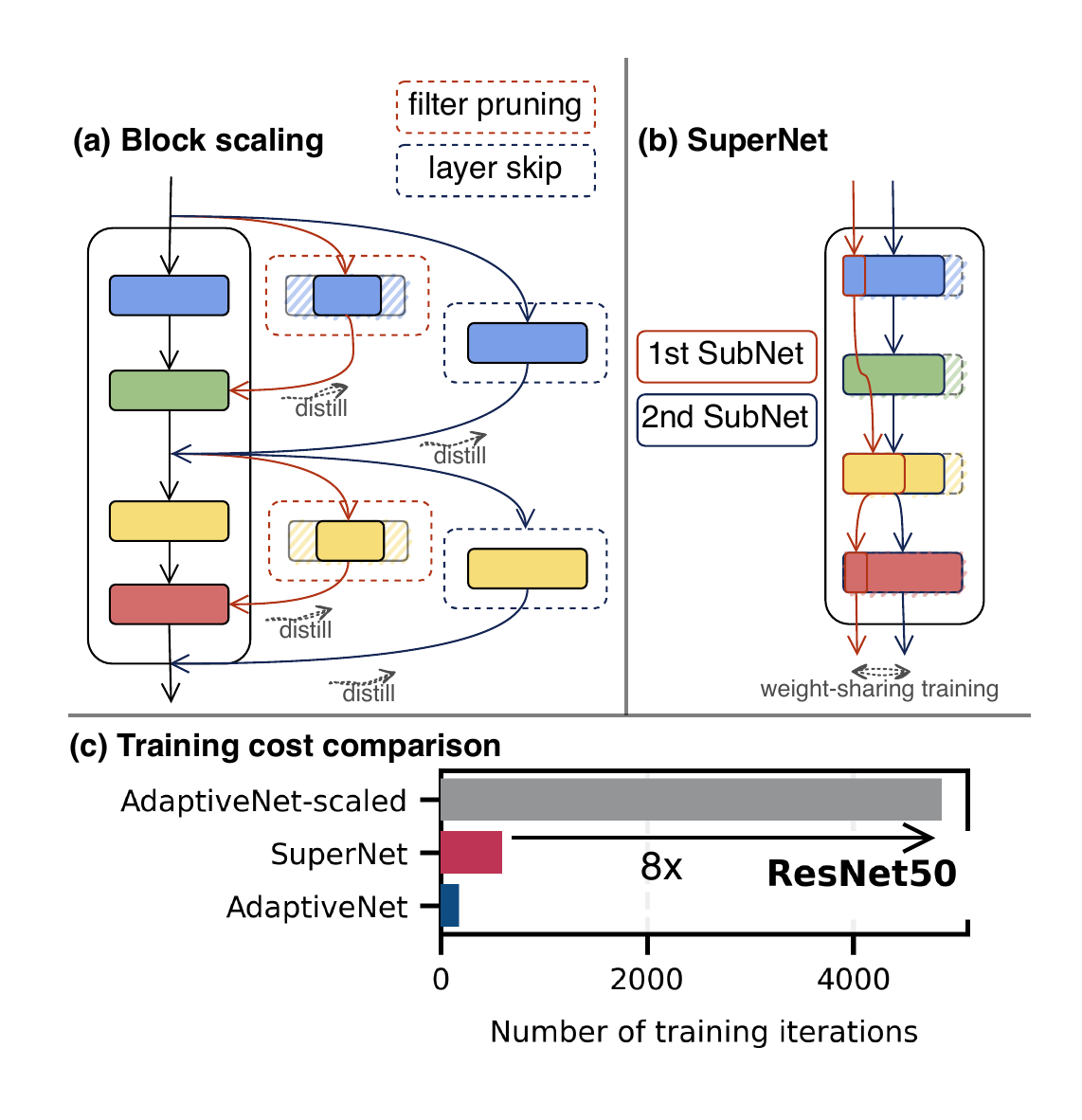}
    \vspace{-35pt}
    \caption{Comparison between block scaling and SuperNet-based model variant generation.}
    \label{fig:figure3}
    \vspace{-10pt}
\end{figure}

\subsection{SuperNet-based Adaptive Model Deployment}
\label{sec2.2}
LegoDNN~\cite{legodnn} and AdaptiveNet~\cite{adaptivenet} adopt a block scaling approach to generate model variants.
As shown in Fig.~\ref{fig:figure3}(a), block scaling constructs multiple block variants, and different model variants are formed by selecting different combinations of these block choices.
Since each new block must be independently constructed and retrained with additional supervision (e.g., block-wise distillation \cite{adaptivenet}), the overhead of memory consumption and training grows linearly with the number of blocks.
Such overhead fundamentally limits scalability, restricting both the number of supported model variants and the granularity of modifications.

To overcome these limitations, we find that the SuperNet \cite{ofa} is a more scalable and efficient model variant generation mechanism.
Unlike block scaling, SuperNet can represent a vast number of SubNets in a weight-sharing manner as shown in Fig. \ref{fig:figure3}(b).
Different SubNets activate distinct subsets of layers or weights within the same SuperNet, enabling the reuse of parameters across variants.
SuperNet training is performed by randomly sampling multiple SubNets per batch, with the total loss aggregated over their outputs. 
This weight-sharing design significantly reduces memory and training overhead, while supporting a much larger and finer-grained variant space.
Scaling AdaptiveNet to match the architectural design space of the SuperNet (excluding input modifications) requires $27\times$ more independent blocks and incurs $27\times$ more training iterations, which is also $8\times$ more than SuperNets, as shown in Fig. \ref{fig:figure3}(c).
Consequently, SuperNet provides a more practical foundation for adaptive model deployment for resource-constrained and dynamic edge environments.

Therefore, we advance SuperNet as the model variant generation method to push the performance frontier of adaptive model deployment.
As shown in Fig.~\ref{fig:figure1}, our experimental results demonstrate the advantages of the SuperNet-based approach compared with existing methods.
Specifically, the SuperNet-based method achieves up to $60.03\%$ latency reduction with the same accuracy and $15.6\%$ absolute accuracy improvement under the same latency constraint compared with AdaptiveNet \cite{adaptivenet} benefiting from its $1,000,000,000\times$ larger design space.

\begin{figure}[t]
    \centering
    \includegraphics[width=\columnwidth]{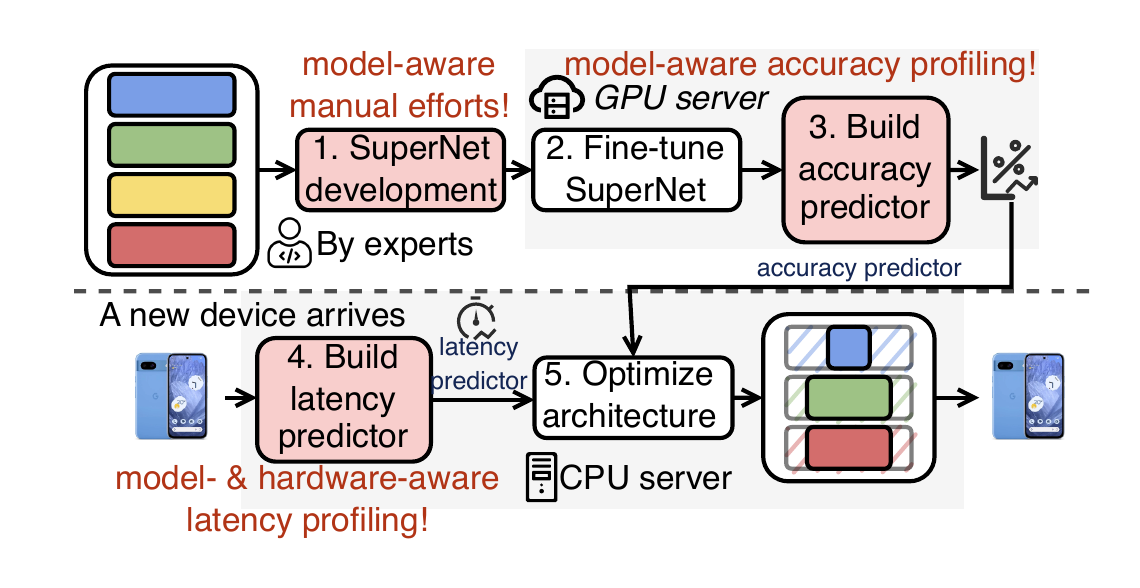}
    \vspace{-30pt}
    \caption{An illustration of the SuperNet workflow. }
    \label{fig:figure4}
    \vspace{-10pt}
\end{figure}

\subsection{Problems in Applying SuperNet}

Despite the significant performance improvements demonstrated by SuperNet-based adaptive deployment, its adoption in real-world systems remains costly.
Fig.~\ref{fig:figure4} illustrates the conventional SuperNet workflow and its inefficiencies.
First, experts manually design and develop a SuperNet (Step 1), requiring significant model-aware expertise.
The SuperNet is then fine-tuned on a GPU server (Step 2) and used to build an accuracy predictor (Step 3).
When a new device arrives, a latency predictor must be built through extensive hardware profiling (Step 4), which incurs high device-specific costs.
Finally, the system combines both predictors to optimize the architecture on an edge server (which can be CPU-only) (Step 5) and deploy the optimized architecture to the target device.
This workflow suffers from two major inefficiencies: heavy reliance on expert manual efforts for SuperNet development and expensive profiling for each new model and device.

\begin{figure*}[t]
    \centering
    \includegraphics[width=\textwidth]{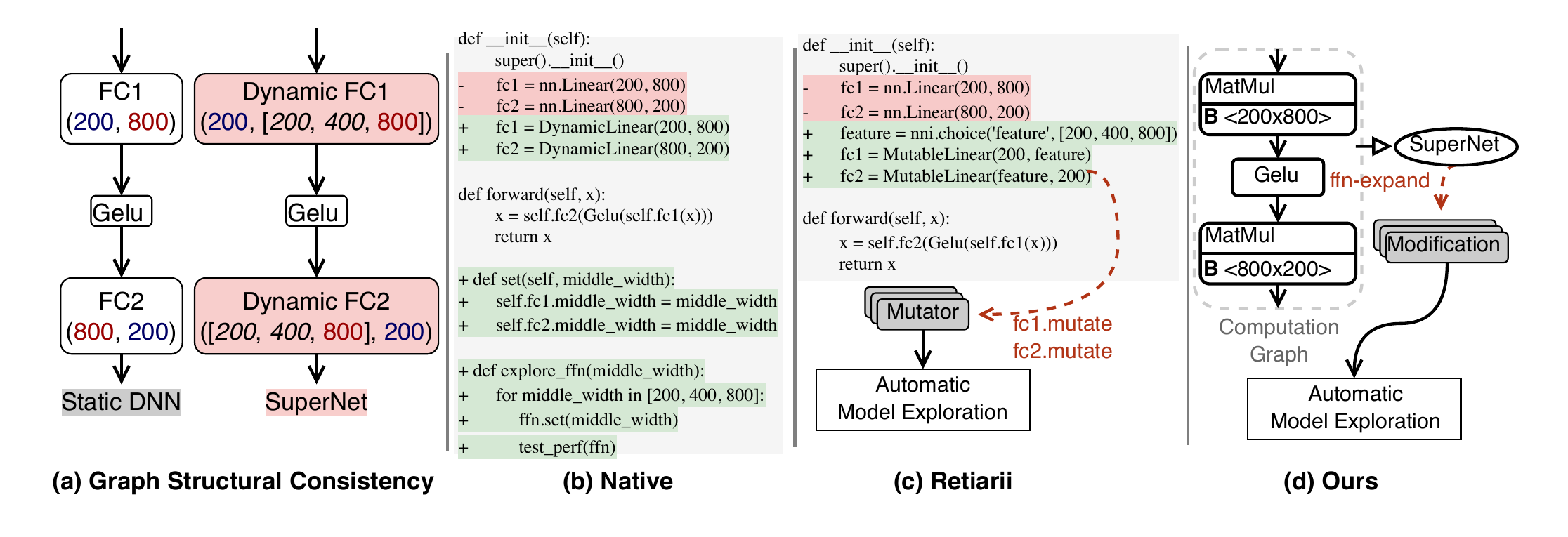}
    \vspace{-35pt}
    \caption{Illustration of the graph structural consistency property and comparison of SuperNet development methods.}
    \label{fig:figure5}
    \vspace{-10pt}
\end{figure*}

On the development side, supporting fine-grained and large-scale variant generation requires complex and manual modifications to the original DNN.
Existing implementations typically extend each layer with the dynamic module and implement its exploration method too as shown in Fig.~\ref{fig:figure5}(b).
This implementation not only inflates the codebase, but also strongly relies on DNN architecture design expertise.
Retiarii~\cite{retiarii} attempts to mitigate the issue via abstractions like \textsc{Mutator} as shown in Fig.~\ref{fig:figure5}(c), which not only keeps the SuperNet code clean, but also reduces repetitive engineering efforts in implementing model exploration required by SuperNet training and architecture optimization through system-level mutator management.
However, developers must still rely heavily on model-aware manual efforts in specifying \textsc{Mutator}s.

\begin{table}[t]
    \centering
    \caption{Profiling costs for deploying SuperNets. Latency profiled on Samsung A54 big cores and accuracy profiled on an NVIDIA RTX 3070 GPU with a batch size of 128.}
    \vspace{-10pt}
    \label{tab:tab1}
    \footnotesize
    \begin{tabular}{lcc}
        \toprule
        \textbf{Model} & \textbf{ResNet50} & \textbf{MobileNetv3} \\
        \midrule
        Number of variants & $2\times10^{13}$ & $7\times10^{22}$ \\
        Latency profiling (s/variant) & $0.2\sim1.75$ & $0.4\sim3.85$ \\
        Accuracy profiling (s/variant) & $31\sim47$ & $31\sim43$ \\
        \bottomrule
    \end{tabular}
    \vspace{-10pt}

\end{table}

\begin{figure}[t]
    \vspace{-15pt}
    \centering
    \includegraphics[width=\columnwidth]{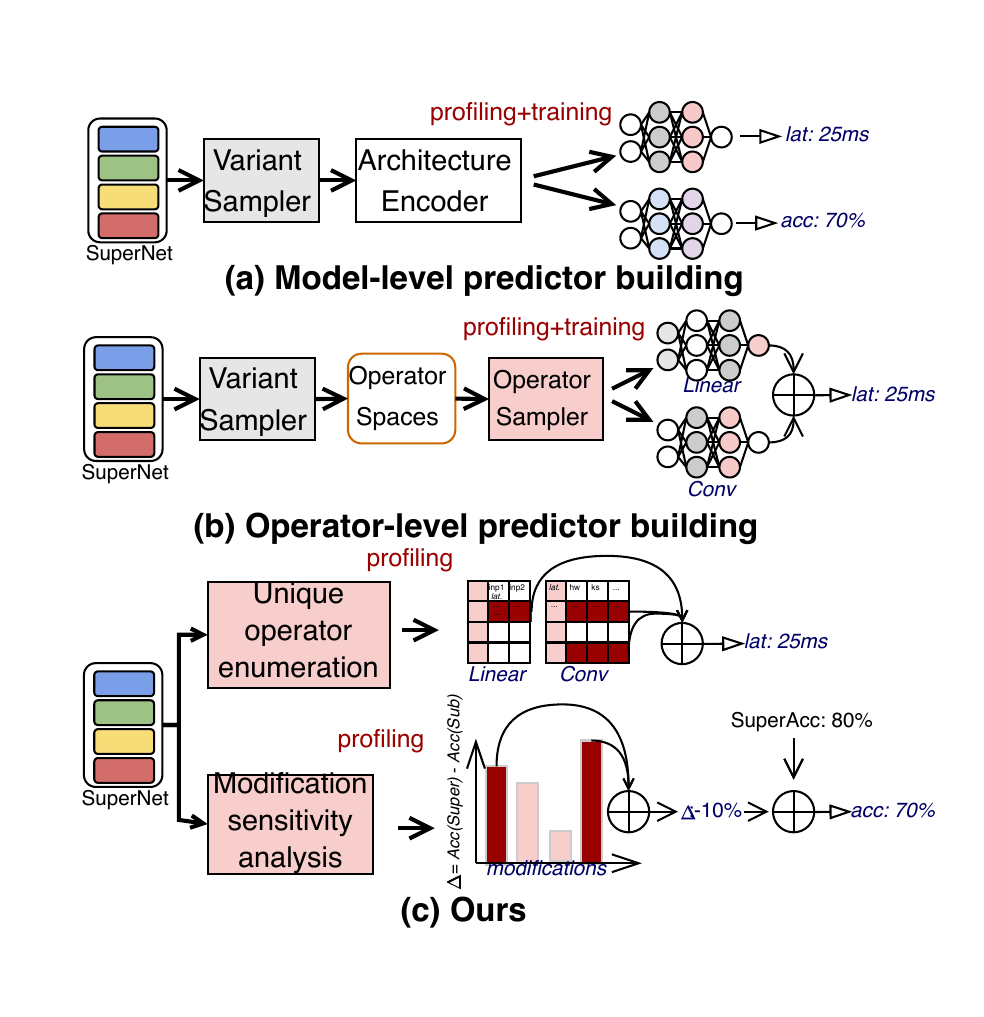}
    \vspace{-40pt}
    \caption{Comparison of predictor building methods.}
    \label{fig:figure6}
    \vspace{-10pt}
\end{figure}

On the profiling side, searching for optimal variants requires performance evaluation across the vast SuperNet design space.
Offline profiling is practical for prior block scaling-based methods with 10,000--20,000 model variants.
However, as illustrated in Table~\ref{tab:tab1}, offline profiling is impractical for a SuperNet with $10^{13}$--$10^{22}$ model variants.
The other way is online profiling, which incurs unacceptable online overhead in the optimization process.
A common alternative is to build predictors trained from partial profiling data.
For example, OFA~\cite{ofa} trains MLP-based predictors that encode subnet architectures and output accuracy or latency estimates as shown in Fig.~\ref{fig:figure6}(a).
However, model-level prediction remains difficult due to the high dimensionality of the architecture space, necessitating substantial profiling to achieve sufficient accuracy.
To reduce cost, operator-level latency predictors have been proposed as shown in Fig.~\ref{fig:figure6}(b), such as nn-Meter~\cite{nnmeter}, which predicts operator-level latencies and aggregates them to estimate end-to-end latency.
LitePred~\cite{litepred} further improves predictor robustness against out-of-distribution sampling.

\begin{figure}[t]
\vspace{-10pt}
        \centering
        \includegraphics[width=\columnwidth]{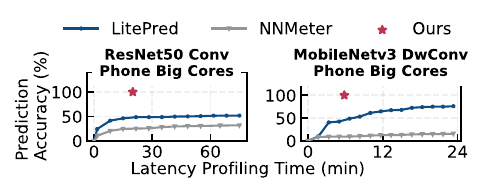}
    \vspace{-30pt}
    \caption{Existing predictor building methods rely on time-consuming profiling for collecting training data.}
    \label{fig:figure7}
    \vspace{-10pt}
\end{figure}

Nevertheless, two fundamental issues persist.
First, predictor accuracy still depends on extensive profiling, often requiring hours of operator measurements.
Second, training predictors introduces its own complexity, demanding careful design choices and hyperparameter tuning.
As shown in Fig.~\ref{fig:figure7}, existing latency or accuracy predictor building methods commonly require hours of data collection to achieve nearly usable predictors.

\subsection{Opportunity}
\label{sec2.4}

\begin{figure*}[t]
    \centering
    \includegraphics[width=0.95\textwidth]{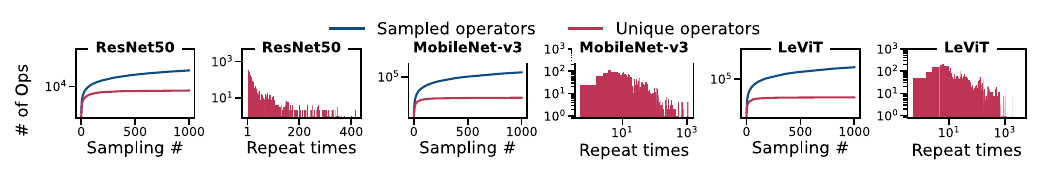}
    \vspace{-15pt}
    \caption{Illustration of cross-SubNet architectural similarity in different models.}
    \label{fig:figure8}
    \vspace{-10pt}
\end{figure*}

To address the development and deployment bottlenecks of SuperNet-based adaptive model deployment, we aim to automate SuperNet construction by isolating users from DNN programs and to devise an efficient method for building latency predictors. However, two challenges arise: (1) how can the framework automatically identify transformable modules and their candidate choices without user intervention? (2) how can accurate predictors be trained in the vast SuperNet design space with limited profiling data?

We identify two exploitable characteristics to address these challenges. \textbf{(1) Graph Structural Consistency:} a SuperNet is derived from a static DNN, where its largest SubNet matches the original architecture.
As a result, the overall computation graph remains consistent with the original DNN (Fig.~\ref{fig:figure5}(a)).
\textbf{(2) Cross-SubNet Architectural Similarity:} analysis of 1,000 randomly sampled SubNets (Fig.~\ref{fig:figure8}) shows that the number of unique operators quickly converges as sampling increases, while many operators appear repeatedly (96--98\% redundancy), often at high frequencies (up to 414--3590 times).
This indicates strong structural redundancy across SubNets, suggesting that the set of unique operators is small and feasible to enumerate.

These two properties create promising opportunities for overcoming existing bottlenecks:

\ul{\textbf{O\#1. Automated SuperNet construction.}}
Leveraging graph structural consistency, static modules can be automatically replaced with dynamic modules during compilation as shown in Fig.~\ref{fig:figure5}(d), eliminating the manual effort required for SuperNet construction.

\ul{\textbf{O\#2. Enumerating unique operators to build learning-free predictors.}} Cross-SubNet similarity enables the enumeration and profiling of all unique operators, which in turn supports efficient construction of lookup-table (LUT)–based predictors as shown in Fig.~\ref{fig:figure6}(c), eliminating the need for costly learning-based models.

%% file: 4_overview.tex
\section{System Overview}

\begin{figure}[t]
    \vspace{-10pt}
    \centering
    \includegraphics[width=\columnwidth]{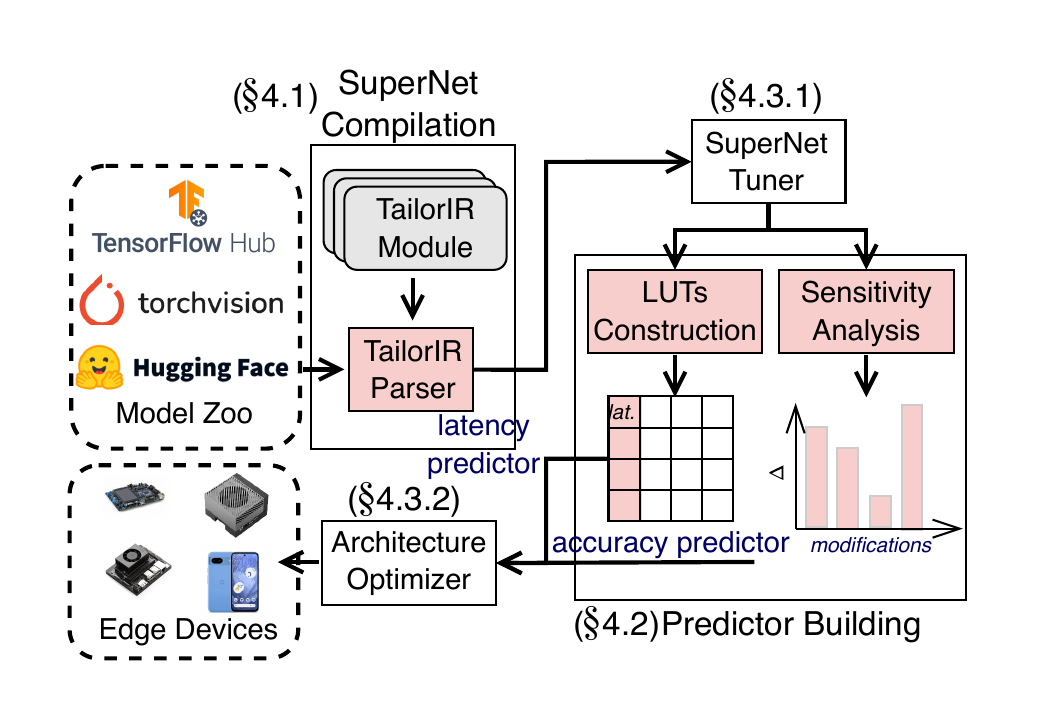}
    \vspace{-30pt}
    \caption{System architecture.}
    \label{fig:figure9}
    \vspace{-10pt}
\end{figure}

AutoTailor advances adaptive model deployment on edge devices by leveraging SuperNets to generate model variants.
It achieves automatic and efficient deployment through two key techniques: (1) a graph compilation method (\S\ref{sec4.1}) that automatically replaces computation graph components without manual coding, and (2) a learning-free predictor construction method (\S\ref{sec4.2}) that rapidly builds latency lookup tables (LUTs) without costly sampling or training.
Given a DNN model and deployment requirements, AutoTailor automatically outputs optimized architectures that deliver state-of-the-art performance on target edge platforms.

\textbf{Challenges and Approaches.}
Despite the opportunities, making automatic and efficient deployment practical requires addressing the following challenges.

\ul{\textit{C\#1. Lack of abstraction.}}
Constructing SuperNets from static DNNs raises a key challenge: how to represent them in a scalable, model-agnostic way.
Retiarii \cite{retiarii} wraps dynamic modules (\textsc{Mutator}) and SuperNets (\textsc{ModelSpace}) around the original DNN (e.g., PyTorch \textit{nn.Module} \cite{torchnnmodule}).
While simple, this approach has two drawbacks: (1) modifications are tightly coupled with the model structure, limiting scalability, and (2) reliance on the original DNN restricts cross-platform support and requires framework-specific versions.

We design TailorIR, a model-agnostic abstraction for automatic SuperNet construction (Section \ref{sec4.1.1}).
SuperNets are organized as computation graphs of TailorIR modules, each managing its architecture and generating executable versions.
This decouples modifications from structure, supports scaling, and ensures traceability.
Similar to Retiarii, TailorIR enables automatic model exploration, but with broader applicability and cleaner integration.

\ul{\textit{C\#2. High engineering effort in dynamic module development.}}
Supporting diverse DNNs requires many dynamic modules, but existing systems like Retiarii make building these libraries labor-intensive.

To this end, TailorIR introduces two optimizations (Section \ref{sec4.1.2}):
(1) bypassing non-variable modules and auto-generating executables with conversion tools;
(2) using templated module designs to maximize reuse across models.

\ul{\textit{C\#3. Inefficiency in enumerating unique operators.}}
Constructing LUTs for learning-free latency prediction requires extracting parameters of unique operators (e.g., kernel size, tensor shapes).
Although their number is limited (Fig. \ref{fig:figure8}), operator architectures are only fixed after model-level modifications, making brute-force sampling infeasible.

To this end, we propose a rapid LUT construction method for our latency predictor (Section \ref{sec4.2.1}) that prunes unnecessary sampling by analyzing independent modifications.
Using TailorIR’s architecture traceability, AutoTailor extracts operator parameters without real inference, greatly speeding up enumeration.

\ul{\textit{C\#4. High accuracy profiling cost.}}
Accuracy predictors usually require full model profiling, which is costly since each candidate must be evaluated on GPUs.
Unlike latency, accuracy cannot be decomposed to the operator level.

To this end, we propose a modification-level accuracy predictor (Section \ref{sec4.2.2}) based on SuperNet modification sensitivities.
Relative accuracy differences between SubNets can be approximated from these sensitivities, allowing efficient prediction without full profiling.

\textbf{System architecture and workflow.}
Fig.~\ref{fig:figure9} shows the system architecture.
The workflow starts with an input DNN, either from a model zoo or a user implementation, represented as a computation graph.
The graph is parsed into a TailorIR-based SuperNet (Section~\ref{sec4.1}), where static modules are replaced with TailorIR modules, and then fine-tuned to preserve SubNet accuracy (Section~\ref{sec4.3.1}).
AutoTailor next builds learning-free predictors (Section~\ref{sec4.2}): the latency predictor profiles unique operators to build device-specific LUTs (Section~\ref{sec4.2.1}), while the accuracy predictor samples SubNets using modification sensitivity to estimate accuracy (Section~\ref{sec4.2.2}).
Finally, given deployment requirements, AutoTailor searches for the best architecture (Section~\ref{sec4.3.2}) and deploys it to the target device.

%% file: 5_design.tex
\section{Detailed Design}

\begin{figure*}[t]
\vspace{-10pt}
    \centering
    \includegraphics[width=\textwidth]{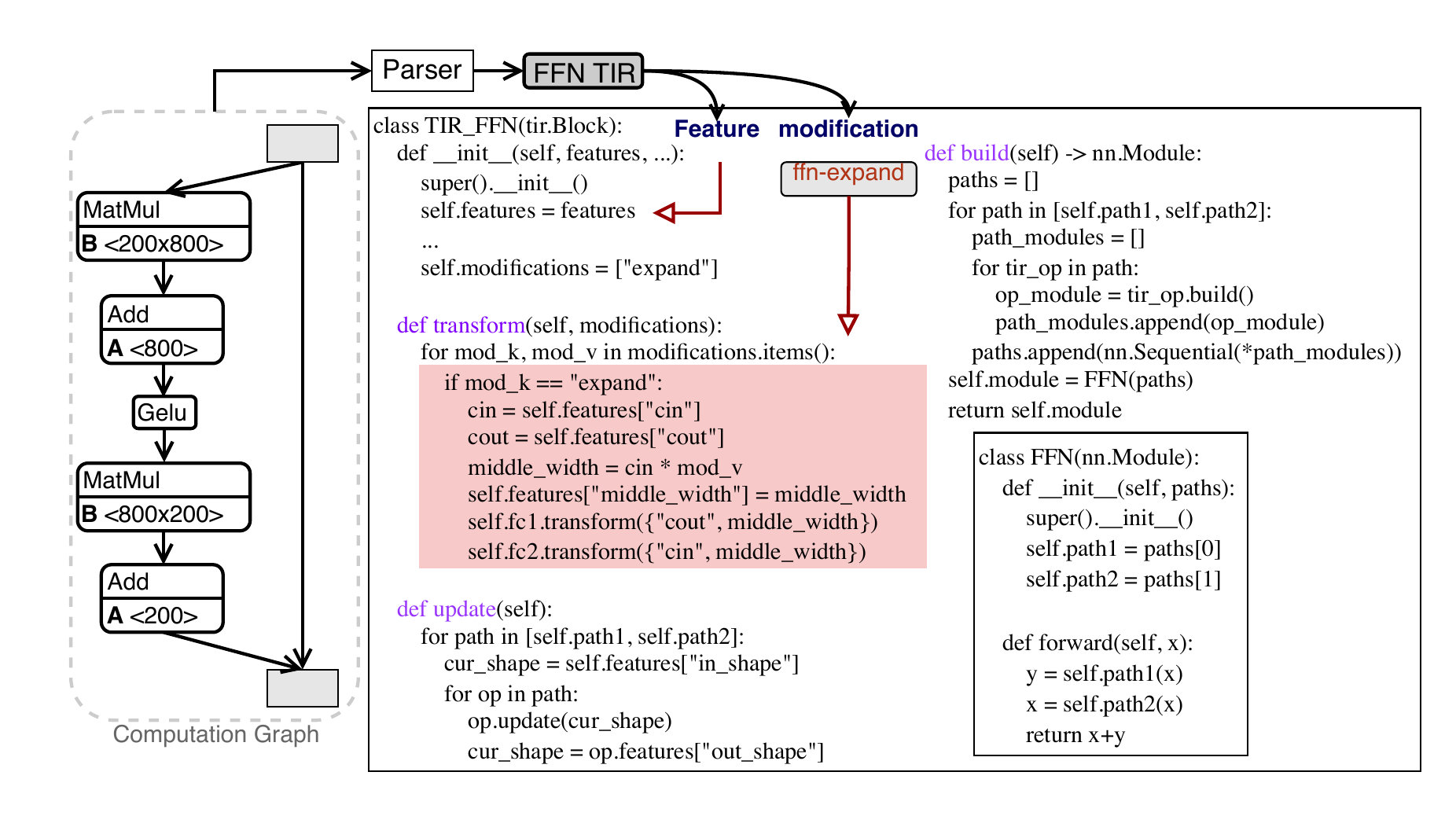}
    \vspace{-35pt}
    \caption{TailorIR overview.}
    \label{fig:figure10}
    \vspace{-10pt}
\end{figure*}

\subsection{Computation Graph-guided Compilation}
\label{sec4.1}

\subsubsection{TailorIR Abstraction.}
\label{sec4.1.1}
Fig.~\ref{fig:figure10} shows the design of TailorIR using an FFN block as an example.
Unlike Retiarii \cite{retiarii}, which mainly wraps PyTorch \textit{nn.Module}, TailorIR is an independent abstraction that directly manages architectures. Specifically:
(1) each TailorIR module maintains a \textit{Feature} describing both meta and active architectures;
(2) the \texttt{transform} function updates features after modifications;
(3) the \texttt{update} function ensures consistency and legality;
(4) the \texttt{build} function generates an executable PyTorch module.
In a SuperNet, operators, blocks, and the full model are all encapsulated as TailorIR modules, each managing its own state to ensure consistent modifications across different levels.

\textbf{Decoupled design.}
TailorIR separates architecture from modifications.
A module’s \textit{Feature} encodes its architecture and is turned into an executable via \texttt{build}, while \texttt{transform} and \texttt{update} handle modifications.
This decoupling improves readability and scalability, as new modifications only require changes to \texttt{transform} without touching \texttt{build}.

\textbf{Architecture traceability.}
A key advantage of TailorIR is architecture traceability, improving module generality and scalability.
Unlike conventional approaches that require manually hardcoding parameters, TailorIR uses a one-time deduction rule in \texttt{update}, so any \texttt{transform} automatically keeps dependent features consistent.

\subsubsection{Module Development Optimizations.}
\label{sec4.1.2}
Automatic construction requires static graphs covered by TailorIR modules and parser rules.
Unsupported operators or architectural patterns require module or parser development; these extensions are reusable across models.
Our module library reduces this effort through two optimizations:

\begin{figure}[t]
\vspace{-10pt}
    \centering
    \includegraphics[width=\columnwidth]{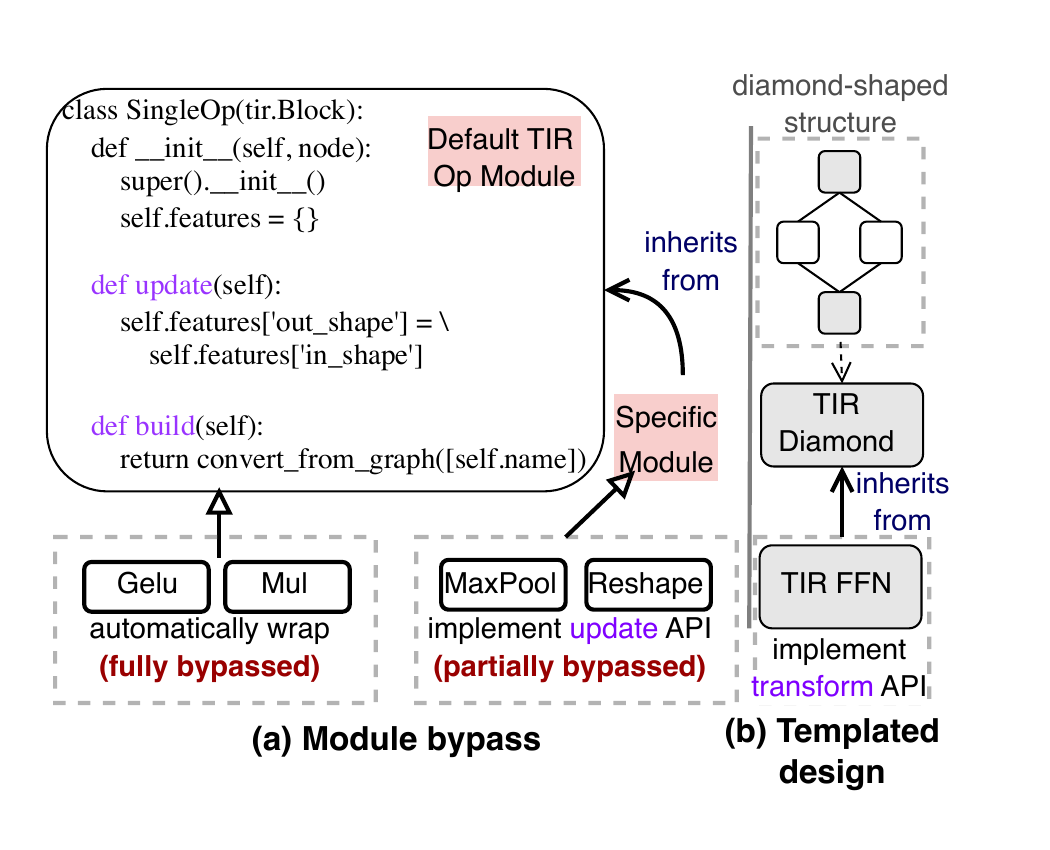}
    \vspace{-35pt}
    \caption{Optimizations in TailorIR module development.}
    \label{fig:figure11}
    \vspace{-10pt}
\end{figure}

\textbf{Static module bypassing.}
As in existing SuperNet methods, not all layers need to be transformed into dynamic modules.
In AutoTailor, since the original program is not directly accessible, we design dedicated TailorIR modules to wrap static layers (Fig.~\ref{fig:figure11}(a)).
These modules maintain a general \textit{Feature} with a default deduction rule in the \textit{update} function to automatically infer attributes from inputs, which works well for non-parameterized and element-wise operators.
During the \texttt{build} stage, model conversion tools (e.g., onnx2torch~\cite{onnx2torch}) generate the corresponding executable PyTorch modules from the computation graph.

\textbf{Templated module design.}
TailorIR enables automatic SuperNet construction and reuses dynamic modules across diverse DNNs.
Although DNNs differ in structure, many share common architectural patterns \cite{modelkeeper,optimus}.
To leverage this, we adopt a templated module design that encourages generic, reusable TailorIR modules.
Using templated modules and the \textit{Path} abstraction (Fig.~\ref{fig:figure11}(b)), TailorIR decouples structure from specific operators, allowing a single design to adapt across networks and reducing redundant engineering.

\subsubsection{Automatic SuperNet Construction.}
\label{sec4.1.3}

We design a bottom-up three-step parsing process that incrementally transforms an input DNN into a TailorIR-based SuperNet.

\textbf{Step 1: Operator parsing.}
The parser traverses the computation graph, mapping each operator to its TailorIR counterpart. Static operators are bypassed (Section~\ref{sec4.1.2}). For multi-output operators, the parser traces output paths until they converge, grouping them into default TailorIR blocks.
\textbf{Step 2: Block parsing.}
The parser then matches subgraphs with templated TailorIR blocks based on topological patterns, replacing default blocks with structured block modules.
\textbf{Step 3: Stage division.}
Finally, the model is partitioned into stages (e.g., by resolution or output width), enabling higher-level modifications such as adjusting stage depth.
For supported graphs, these steps construct a SuperNet without manual model rewriting.

\subsection{Learning-free Predictor Building}
\label{sec4.2}
We propose a learning-free predictor that avoids the overhead and errors of learning-based methods. For latency, AutoTailor profiles only unique operators to build operator-level LUTs, keeping costs low and eliminating generalization errors. To identify these operators efficiently, we prune the SubNet enumeration space using modification dependencies. We extend this idea to accuracy with \textit{modification sensitivity}, which measures the impact of architectural changes and enables efficient modification-level prediction.

\begin{figure}[t]
    \centering
    \includegraphics[width=\columnwidth]{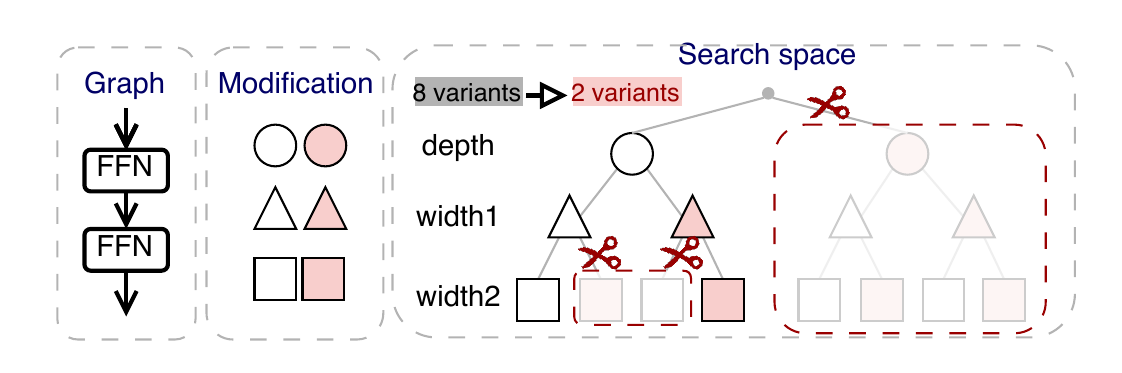}
    \vspace{-25pt}
    \caption{An example of space pruning.}
    \label{fig:figure12}
    \vspace{-10pt}
\end{figure}

\subsubsection{Operator-level LUT-based Latency Predictor.}
\label{sec4.2.1}
AutoTailor accelerates LUT construction by pruning redundant operator enumerations through analysis of modification dependencies.

\textbf{Pruning the enumeration space.}
As shown in Fig.~\ref{fig:figure12}, depth modifications determine only the number of stacked blocks within a stage. 
Because these blocks are structurally identical, the corresponding operator features remain unchanged regardless of depth, making such modifications independent.
Similarly, modifications confined to a specific stage or block are also independent of other choices.
By exploiting such independence, AutoTailor prunes redundant operator combinations from the enumeration space, thereby avoiding unnecessary profiling.

\begin{table}[t]\centering
\caption{Extraction cost of unique operators. In the normal setting, the cost is estimated by multiplying the number of profiling runs by the average cost of a single profiling run.}
    \vspace{-10pt}
    \label{tab:extractcost}
    \footnotesize
    \resizebox{\columnwidth}{!}{
\footnotesize
\begin{tabular}{|c|c|c|c|}
\hline
\multirow{2}{*}{\textbf{Models}} & \textbf{normal} & \textbf{+ space pruning} & \textbf{+ TailorIR} \\
& (GPU hours) & (GPU hours) & (CPU minutes) \\
\hline
ResNet50 & 1.25E+09 & 10.8 & 11.8 \\
\hline
MobileNetv2 & 872000 & 0.0382 & 2.2 \\
\hline
ConvNext & 8.08978E+15 & 0.0259 & 4.8 \\
\hline
LeViT & 1.02E+41 & 57.44 & 46.5 \\
\hline
\end{tabular}
}

    \vspace{-10pt}
\end{table}

Beyond the large enumeration space, the cost of extracting operators from a single SubNet is also non-negligible.
Tensor shapes are essential operator features for constructing LUTs, yet obtaining them typically requires inference of the entire SubNet due to the dynamic graph design in existing deep learning frameworks.
AutoTailor alleviates this overhead by leveraging TailorIR’s architectural traceability, which automatically infers the tensor shape of each module without full SubNet execution.
As shown in Tab. \ref{tab:extractcost}, this optimization makes model-aware enumeration both efficient and practical.

\textbf{Latency predictor building.}
Once the LUT is built, each device profiles its unique operators. For a given SubNet, AutoTailor extracts each operator’s \textit{Feature}, looks up its latency in the device-specific LUT, and sums them to get the model’s total latency. This design ensures accuracy and easy transfer across devices.
Operator fusion \cite{dnnfusion} merges multiple operators into one kernel, breaking the assumption that total latency equals the sum of individual latencies \cite{nnmeter}. To handle this, AutoTailor extends TailorIR to support fusion: multiple operators can be combined into a single TailorIR module based on user-defined fusion rules. During LUT construction, each fused operator is built and profiled as a single module, so predictors capture the true execution cost. Section~\ref{sec6.4} shows that this fusion-aware approach achieves accurate latency prediction.

\subsubsection{Modification-level Accuracy Predictor.}
\label{sec4.2.2}
\begin{figure}[t]
    \centering
    \includegraphics[width=\columnwidth]{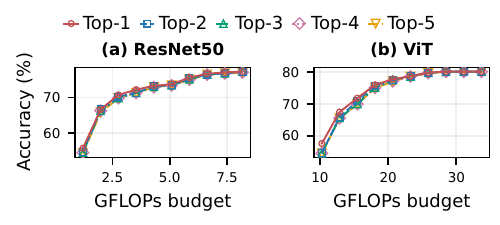}
    \vspace{-30pt}
    \caption{Accuracy of top-ranked SubNets across FLOPs budgets. Top-$k$ denotes the minimum measured accuracy among the $k$ highest-ranked candidates in each search.}
    \label{fig:additive-selection-topk}
    \vspace{-15pt}
\end{figure}

The primary goal of our accuracy predictor is to identify high-performing SubNets through approximate ranking rather than precisely predict their absolute accuracy.
Fig.~\ref{fig:additive-selection-topk} evaluates the effectiveness of this design choice.
Across the evaluated budgets, Top-5 accuracy is only 0.64\% lower than Top-1 accuracy on average for ResNet50 and 0.92\% lower for ViT.
These results suggest that approximate ranking is sufficient to identify competitive SubNets, reducing the need for highly precise accuracy estimates.

To enable efficient ranking, we leverage the observation that all SubNets inherit weights from the same trained SuperNet and differ only in their architectural modifications.
This shared foundation allows us to approximate the accuracy impact of combined modifications using the sensitivity of individual modifications, avoiding exhaustive evaluation of their combinations.
Specifically, let $A_0$ denote the accuracy of the unmodified SuperNet and $A(S)$ the measured accuracy after applying a set of modifications $S$.
AutoTailor profiles each modification $m$ individually on the training dataset and estimates the accuracy of a candidate SubNet as
\begin{equation}
    \Delta_m=A_0-A(\{m\}),\qquad
    \widehat{A}(S)=A_0-\sum_{m\in S}\Delta_m.
\end{equation}
Here, $\Delta_m$ captures the accuracy sensitivity of modification $m$, while summing individual sensitivities approximates their combined impact without explicitly modeling interactions.
This additive approximation enables efficient, architecture-aware SubNet ranking without the overhead of evaluating every candidate.

\begin{figure}[t]
    \centering
    \vspace{-10pt}
    \includegraphics[width=\columnwidth]{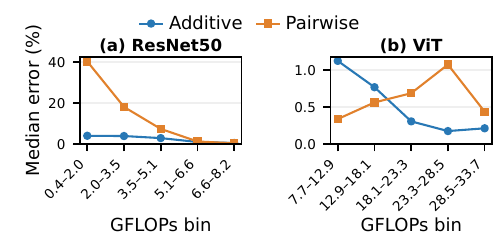}
    \vspace{-30pt}
    \caption{Median accuracy error of additive and pairwise predictors across FLOPs bins, with 100 SubNets per bin.}
    \label{fig:additive-flops-bins}
    \vspace{-10pt}
\end{figure}

Fig.~\ref{fig:additive-flops-bins} examines the effectiveness of our additive approximation across different computational scales.
For each model, we sample 100 distinct SubNets from each of five FLOPs bins.
For ResNet50, the median prediction error decreases from 4.03\% to 0.49\% across bins, while for ViT, it ranges from 0.18\% to 1.13\%.

A limitation of the additive approximation is that it ignores interactions among architectural modifications.
To assess whether modeling such interactions improves prediction accuracy, we extend the predictor with a pairwise correction term, $A(\{a,b\})-A(\{a\})-A(\{b\})+A_0$, for each co-occurring pair of modifications.
Surprisingly, explicitly modeling pairwise interactions does not consistently improve prediction accuracy.
For ResNet50, the median errors range from 40.39\% to 0.35\% across bins, while for ViT, they range from 0.34\% to 1.08\%, despite requiring 33$\times$ and 14.8$\times$ more profiling configurations, respectively.
This result suggests that pairwise corrections do not necessarily capture the combined effects of architectural modifications, as higher-order interactions remain unmodeled and individual correction terms may introduce additional estimation errors.
Overall, these findings support additive sensitivity as a practical approximation that balances prediction accuracy and profiling overhead.

\subsection{Automatic Model Exploration}
\label{sec4.3}
Model exploration samples SubNets from the search space and is crucial for (1) fine-tuning the SuperNet and (2) architecture optimization, where the best candidates are selected using the fine-tuned SuperNet and predictors.
AutoTailor automates this by collecting modification abstractions from each TailorIR module; users only specify modifications and candidate choices in a simple configuration file:

\begin{lstlisting}[style=toml]
title = "Example"
[arch]
blocks = ["FFNBlock"]
[var.global_vars]
resolution = [128, 160, 192, 224]
[var.stage_vars]
reduce_depth = [-3, -2, -1, 0]
[var.block_vars]
FFNBlock.expand_ratio = [2, 3, 4]
\end{lstlisting}

\subsubsection{SuperNet Fine-tuning}
\label{sec4.3.1}
Fine-tuning is necessary to ensure that modified SuperNets preserve competitive accuracy.  
AutoTailor incorporates a \textit{Tuner} component to execute this process, providing seamless integration with existing SuperNet-based model exploration strategies \cite{compofa,elasticvit}.  
Given a TailorIR-represented SuperNet, the Tuner samples SubNets by applying modifications and fine-tunes them accordingly.

\subsubsection{Architecture Optimization}
\label{sec4.3.2}
The \textit{Optimizer} in AutoTailor efficiently searches for the best architecture to deploy fine-tuned SuperNets on edge devices using the constructed predictors. To handle this complex task, AutoTailor uses genetic algorithms \cite{ofa,elasticvit} and accelerates the search with a beam search-based initialization for large design spaces.

%% file: 7_impl.tex
\section{Implementation}
\label{sec5}
We implement AutoTailor on PyTorch \cite{pytorch} in 12500+ LoC of Python, including 4700+ LoC for the library of TailorIR modules to support all SuperNets mentioned in this paper.
We use ONNX \cite{onnx} and ONNX Graph Surgeon \cite{gs} tools for parsing the model.
The tuner component is implemented on PyTorch DistributedDataParallel (DDP) for supporting distributed SuperNet tuning.
The optimized architecture will be translated into NCNN executable format by the PNNX \cite{pnnx} tool for deploying on mobile phones.

%% file: 8_evaluation.tex
\section{Evaluation}

\subsection{Experimental Setup}

\textbf{Testbed.}
We evaluate AutoTailor on four edge setups: a Samsung A54 smartphone (Exynos 1380) using its Big- and Little-cores, and NVIDIA Jetson Orin Nano and AGX Orin GPUs. AutoTailor runs on a local server with NVIDIA 3070 GPUs and Intel Xeon Gold 6230R CPUs, offloading SuperNet tuning to the cloud. NCNN \cite{ncnn} is used for on-device inference (FP16 precision), except for ViT on Jetson devices, where PyTorch \cite{pytorch} performs better.

\textbf{Models and datasets.}
Tab.~\ref{tab3} lists the SuperNets and their original models used in our experiments. All SuperNets support width, depth, and input modifications. We use ImageNet1K \cite{im1k} for both fine-tuning and evaluating adapted models. Following CompOFA \cite{compofa}, each SuperNet is fine-tuned for 600 iterations (150 epochs) with 4 SubNets sampled per batch.

\begin{figure*}[t]
    \centering
    \includegraphics[width=\textwidth]{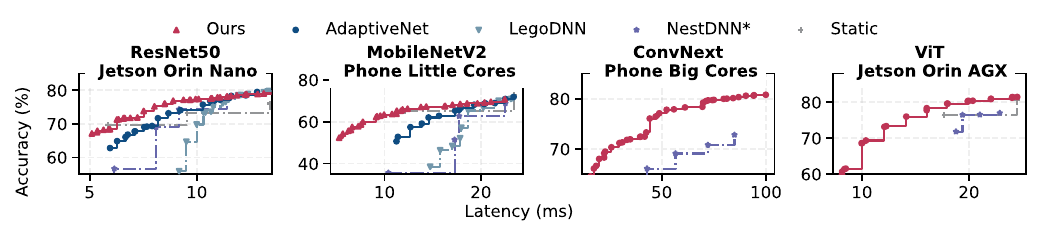}
    \vspace{-30pt}
    \caption{Performance comparison of hardware-aware adaptive model deployment.}
    \label{fig:figure13}
    \vspace{-10pt}
\end{figure*}

\begin{figure*}[t]
    \centering
    \includegraphics[width=\textwidth]{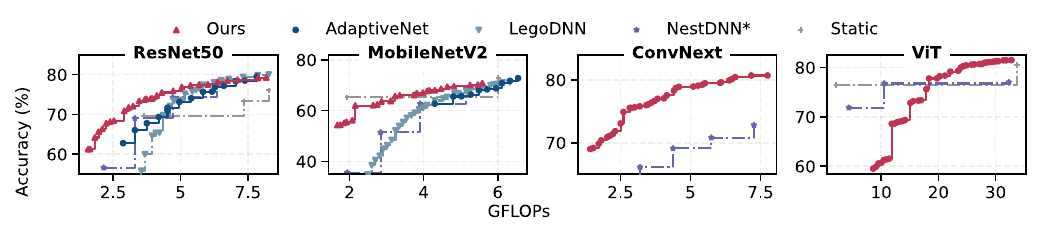}
    \vspace{-30pt}
    \caption{Performance comparison of hardware-agnostic adaptive model deployment.}
    \label{fig:figure14}
    \vspace{-15pt}
\end{figure*}

\begin{table}[b]
    \vspace{-10pt}
    \centering
    \caption{Illustration of SuperNets in this paper.}
    \vspace{-10pt}
    \label{tab3}
    \footnotesize
    \begin{tabular}{lcc}
        \toprule
        \textbf{Original DNN} & \textbf{\# of Modifications}& \textbf{\# of Variants} \\
        \midrule
        ResNet50 \cite{resnet} & 100 & $2\times10^{13}$\\
        \midrule
        MobileNetv2 \cite{mobilenetv2} & 85 & $8\times10^{10}$\\
        \midrule
        MobileNetv3 \cite{mobilenetv3} & 160 & $7\times10^{22}$ \\
        \midrule
        ProxylessNet \cite{proxylessnet} & 166 & $6\times10^{23}$ \\
        \midrule
        ConvNext-Tiny \cite{cnnet} & 163 & $2\times10^{23}$ \\
        \midrule
        ViT-Base \cite{vit} & 103 & $1\times10^{12}$\\
        \midrule
        LeViT \cite{levit} & 324 & $1\times10^{45}$\\
        \bottomrule
    \end{tabular}

\end{table}

\subsection{Adaptive Deployment Performance}
We first evaluate AutoTailor’s adaptive deployment performance by searching for optimal architectures across different models, edge devices, and requirements.
For each case, we generate 40–50 latency constraints and report only the Pareto frontier of the resulting latency–accuracy trade-offs.
Our adaptation uses our proposed learning-free predictors for providing both latency and accuracy guidance in the architecture optimization.
The hyperparameter settings of the genetic search algorithm used in architecture optimization follow those in OFA \cite{ofa}.

\textbf{Baselines.}
We compare against AdaptiveNet \cite{adaptivenet} and LegoDNN \cite{legodnn} using their official implementations. AdaptiveNet uses its released weights, while LegoDNN trains new blocks on pre-trained DNNs from TIMM \cite{timm}. For NestDNN, whose code is unavailable, we approximate its performance using Torch-Pruning \cite{depgraph} and denote it as NestDNN*. Pruned variants are fine-tuned for 60 epochs on ResNet50 and MobileNetV2, and 120 epochs on ConvNeXt and ViT. Since AdaptiveNet and LegoDNN do not support Transformers, we compare them only against NestDNN* for these models. We also report static DNN performance for reference.

\textbf{Results.}
Due to space constraints, we select one representative model per runtime, pairing larger models with stronger devices and smaller models with weaker ones.
As shown in Fig.~\ref{fig:figure13}, AutoTailor supports a wider range of latency constraints and consistently delivers higher accuracy under 92\% of the evaluated latency conditions than existing methods.

On ResNet50, AutoTailor achieves an average accuracy gain of $2.27\%$ and a $10.37\%$ latency reduction over AdaptiveNet.
On MobileNetV2, the improvements are larger, with $4.12\%$ higher accuracy and $21.48\%$ lower latency.
These gains are most pronounced on mobile phones, which have fewer parallel resources than Jetson GPUs and thus require more compact models. AutoTailor generates such compact variants more effectively than existing methods.
In contrast, LegoDNN and NestDNN* perform poorly under tight latency budgets, especially on weaker devices, due to their lack of depth modifications.
For larger latency budgets (8\%), AutoTailor is slightly less competitive, which we attribute to limited SuperNet training. We expect this gap can be closed with stronger tuning strategies, such as OFA’s progressive training \cite{ofa} ($3\times$ more iterations) or AlphaNet’s self-distillation \cite{alphanet} ($1.25\times$ more iterations).

On Transformer-style models, AutoTailor improves accuracy by $5.12\%$ and reduces latency by $29.62\%$ on ConvNeXt, and achieves $3.36\%$ higher accuracy with $15.04\%$ lower latency on Vision Transformer compared with NestDNN*.
These results demonstrate AutoTailor’s effectiveness on advanced architectures and its potential for emerging systems such as vision foundation models \cite{aria} and vision–language models \cite{llava}, where ConvNeXt and Vision Transformer serve as common image encoders.

We also report hardware-agnostic results by using FLOPs instead of latency (Fig.~\ref{fig:figure14}).
Unlike hardware-aware results, the FLOPs-guided ones show different trends, underscoring the need for accurate hardware-specific predictors.
Across most FLOPs constraints, AutoTailor still achieves state-of-the-art performance.
For Vision Transformer, width pruning performs well, likely due to its higher architectural complexity, which may benefit from more SuperNet fine-tuning.

\begin{figure}[t]
    \centering
    \includegraphics[width=\columnwidth]{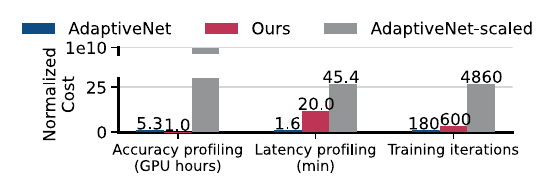}
    \vspace{-30pt}
    \caption{Overhead analysis of ResNet50 adaptive model on phone big cores.}
    \label{fig:figure15}
    \vspace{-15pt}
\end{figure}

\textbf{Overhead analysis.}
Next, we analyze the overhead of applying SuperNets.
Fig.~\ref{fig:figure15} compares the extra cost from AutoTailor’s larger design space with AdaptiveNet and a scaled version (AdaptiveNet-scaled) that matches our search space.
The official AdaptiveNet validates accuracy on only a data subset, which reduces reliability.
For fairness, we validate all 10,609 variants and include this cost in our analysis.

Our sensitivity-based predictor needs only one hour of profiling.
Latency profiling takes $12.5\times$ longer than AdaptiveNet but only 20 minutes in absolute time, which is practical.
AdaptiveNet requires 60 training iterations with 3 sampled blocks per batch, while our method incurs $3.3\times$ more iterations to fine-tune SuperNets.
The AdaptiveNet-scaled comparison highlights our better scalability, even without input modifications that would further raise its overhead.

\subsection{Effectiveness of Computation-graph Guided Compilation}

\begin{figure}[t]
    \centering
    \includegraphics[width=\columnwidth,height=0.340474\columnwidth]{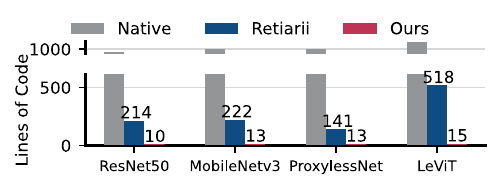}
    \vspace{-25pt}
    \caption{Comparison of LoC in constructing SuperNets.}
    \label{fig:figure16}
    \vspace{-15pt}
\end{figure}

\textbf{Engineering Effort Reduction.}
We quantify the engineering effort required for SuperNet construction by measuring lines of code (LoC) changes relative to the original DNN using \texttt{git diff}. We evaluate four SuperNets (ResNet50, MobileNetV3, ProxylessNet, and LeViT), using their official OFA~\cite{ofa} and ElasticViT~\cite{elasticvit} implementations as native baselines.
The Retiarii baseline is implemented by us using their open-source framework, NNI \cite{nni}.
With AutoTailor, users only need to provide a simple configuration file specifying modifications (Section \ref{sec4.3}).
As shown in Fig.~\ref{fig:figure16}, AutoTailor reduces LoC by 10.8--34.5$\times$ compared with Retiarii, thanks to automatic graph compilation.
While Retiarii eases model exploration, it still requires manual replacement of static modules with mutable ones, which demands both detailed architecture knowledge and significant engineering.
In contrast, AutoTailor eliminates such model-aware coding by letting users define SuperNets through concise, easy-to-write configuration files.

\begin{figure}[t]
    \centering
    \includegraphics[width=\columnwidth,height=0.340474\columnwidth]{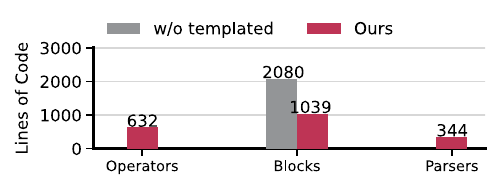}
    \vspace{-30pt}
    \caption{Analysis of TailorIR Modules.}
    \label{fig:figure17}
    \vspace{-15pt}
\end{figure}

\textbf{Effectiveness of Development Optimization.}
Fig.~\ref{fig:figure17} analyzes the LoC distribution of core functions, including \texttt{build}, \texttt{transform}, \texttt{update}, and \texttt{init} functions, across TailorIR modules developed for the SuperNets of Tab.~\ref{tab3}.
About 51\% of the effort goes into 12 templated blocks, with nearly 50\% saved through LoC reuse. Operator development takes 632 LoC for 21 operators (10 partially bypassed), out of 28 total operators (7 fully bypassed). Parser development accounts for about 20\% of the effort, as it maps sub-graphs to block templates. To simplify and avoid conflicts, users can specify block templates directly in the config file (Sec.~\ref{sec4.3}).

\subsection{Effectiveness of Learning-free Predictors}
\label{sec6.4}

\textbf{Baselines.}
For latency prediction, we compare AutoTailor with LitePred \cite{litepred}, extending both to support additional operators. Each SuperNet uses 2000 sampled SubNets to build predictors. For accuracy prediction, we compare with OFA’s MLP-based method \cite{ofa}.

\textbf{Metrics.}
For latency prediction, we report error rates at the 10\% level on 1000 randomly sampled SubNets following~\cite{nnmeter,litepred}. For accuracy prediction, we use top-5 ranking accuracy: 100 groups of 200 candidates are ranked by predicted accuracy. We adopt ranking accuracy instead of absolute values to better match the optimization process, which selects the best SubNets within each group.

\begin{figure}[t]
    \centering
    \includegraphics[width=\columnwidth]{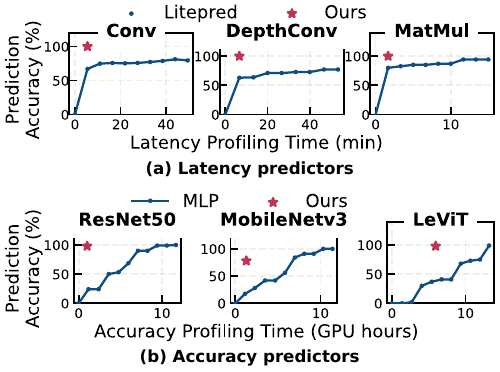}
    \vspace{-20pt}
    \caption{Profiling cost of (a) latency predictors for LeViT operators and (b) accuracy predictors across models.}
    \label{fig:figure18}
    \vspace{-10pt}
\end{figure}

\begin{figure}[t]
    \centering
    \includegraphics[width=\columnwidth]{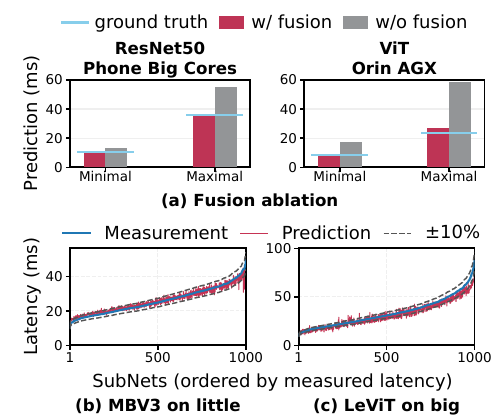}
    \vspace{-20pt}
    \caption{Model-level latency prediction: (a) fusion ablation; (b,c) measured and predicted latency for 1000 SubNets per setting on Samsung A54. Dashed lines mark $\pm10\%$ of measured latency.}
    \label{fig:figure19}
    \vspace{-10pt}
\end{figure}

\textbf{Predictor profiling cost reduction.}
As shown in Fig.~\ref{fig:figure7}, existing learning-based predictors struggle with accuracy and efficiency. Our learning-free method profiles all unique operators for precise latency estimation, taking only 20 minutes for ResNet50’s convolutions and 40 minutes for MobileNetV3’s depth-wise convolutions, which are their main latency contributors.
As shown in Tab.~\ref{tab3}, LeViT has the most model variants, making predictor construction challenging. Fig.~\ref{fig:figure18}(a) compares our method with LitePred \cite{litepred} on three major operators. Our method reduces latency-predictor profiling time by 8--9$\times$ compared with the largest LitePred budgets shown, while achieving higher prediction accuracy.
For accuracy prediction, Fig.~\ref{fig:figure18}(b) shows that our method reduces predictor construction time by 2.2--11.5$\times$ compared to the full MLP profiling runs shown while achieving 78--98\% top-5 ranking accuracy.
Our method offers an efficient alternative for rapidly constructing predictors.

\textbf{Model-level latency prediction.}
We compare predicted and measured latencies of the smallest and largest SubNets, with and without fusion.
As shown in Fig.~\ref{fig:figure19}(a), ignoring fusion causes errors up to $1.47\times$, especially in larger SubNets with more fused operators.
While perfect accuracy is difficult, our fusion-aware method keeps deviations within an acceptable range.
We further choose MobileNetV3 and LeViT to evaluate memory-intensive execution.
We evaluate predicted and measured model-level latency for 1000 randomly sampled SubNets of each model on the Samsung A54's big and little CPU cores.
Fig.~\ref{fig:figure19}(b,c) shows that most predictions lie within the $\pm10\%$ bounds in both settings.

\begin{figure}[t]
    \centering
    \includegraphics[width=\columnwidth]{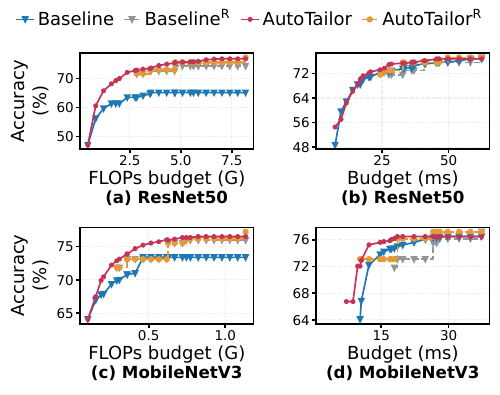}
    \vspace{-25pt}
    \caption{Ablation study on computation-quality trade-offs. Baselines: MLP in (a,c) and MLP+LitePred in (b,d). $\mathrm{R}$ denotes the restricted design space.}
    \label{fig:ablation-study}
    \vspace{-10pt}
\end{figure}

\subsection{Ablation Study}
\label{sec:ablation}
To evaluate how AutoTailor's design space and predictors contribute to better computation-quality trade-offs, we ablate both components in Fig.~\ref{fig:ablation-study}.
The full space provides 44.9\% and 22.5\% wider observed FLOPs ranges than the five restricted presets (R) for ResNet50 and MobileNetV3, respectively.
Within the full space, we compare AutoTailor's accuracy predictor with MLP in (a,c), and its accuracy-latency predictor pair with MLP+LitePred in (b,d).
Baseline$^{\mathrm{R}}$ often outperforms Baseline, suggesting that searching the larger space amplifies baseline ranking errors and makes reliable predictors essential to realizing its benefits.
Our predictors improve accuracy by 10.43\%/2.90\% at matched FLOPs and 11.36\%/3.14\% at matched summed A54 operator latency for ResNet50/MobileNetV3, respectively.

\subsection{Generality Case Studies}
\label{sec:generality}
\begin{figure}[t]
    \centering
    \includegraphics[width=\columnwidth]{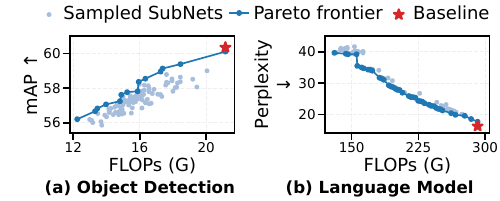}
    \vspace{-20pt}
    \caption{Computation–quality trade-offs for object detection and language modeling.}
    \label{fig:generality}
    \vspace{-10pt}
\end{figure}

To evaluate the generality of AutoTailor, we extend the framework to support object detection and autoregressive language modeling.

\textbf{Setup.}
(1) Object detection. We use the HardHat~\cite{hardhat} dataset, split into 4000 training and 1000 validation images.
We construct the detector SuperNet by combining a ResNet-style TailorIR backbone with a YOLOv8~\cite{yolov8} detector.
Only the backbone is transformable, while the neck and head architectures are fixed.
We jointly train the backbone, neck, and heads for 70 additional epochs at a resolution of ($640\times640$).
Each batch trains the largest and smallest SubNets together with two randomly sampled SubNets.
We measure detection quality using mean average precision (mAP)~\cite{pascal-voc}.
(2) Autoregressive language modeling. We train a GPT-2~\cite{gpt2} SuperNet with 4--12 Transformer blocks and feed-forward expansion ratios of 2--4 on WikiText-103~\cite{wikitext103} for 20000 iterations following nanoGPT~\cite{nanogpt}.
We measure language-model quality using perplexity~\cite{bengio2003}.
For both tasks, we select SubNets using CompOFA~\cite{compofa} presets and FLOPs-stratified random sampling.
The baseline for each task is the corresponding full-size model fine-tuned on the same dataset using the same training procedure.

\textbf{Results.}
Fig.~\ref{fig:generality} shows the computation-quality trade-offs for both tasks.
(1) Object detection. The Pareto frontier spans 12.24--21.15 GFLOPs and 56.21--60.10\% mAP.
(2) Autoregressive language modeling. The GPT-2 SubNets span 130.6--291.6 GFLOPs per sequence and achieve perplexities of 17.66--41.62.
Object detection reuses existing parser support and block templates with zero additional LoC; GPT-2 requires 134 LoC for parser support and 261 LoC for templated blocks (395 LoC total).
Together, these results demonstrate AutoTailor's extensibility across diverse model architectures and tasks with modest engineering effort.

%% file: 9_relatedwork.tex
\section{Related Work}

\textbf{SuperNet Architectures and Training.}
Weight-sharing SuperNets support adaptive architectures across vision~\cite{ofa,usnet,autoformer,elasticvit,superdetection}, speech~\cite{superasr}, and language tasks~\cite{hat,dynabert,flextron,llamaflex}.
Industrial-scale foundation models also adopt this paradigm, including Google's Gemma 3n~\cite{matformer,gemma3n} and NVIDIA's Nemotron Elastic~\cite{nemotronelastic}.
Their high training cost motivates algorithmic optimizations, including search-space restriction~\cite{greedynas,compofa,deps,bignas} and knowledge distillation~\cite{attentivenas,alphanet,magicat,elasticvit,momentumnas}, and system-level parallelization~\cite{naspipe,pathparallel}.

\textbf{Adaptive Inference Systems.}
Using model variants as a control knob is a widely adopted strategy for meeting diverse edge deployment requirements~\cite{nestdnn,legodnn,adaptivenet} and adapting to fluctuating cloud serving workloads~\cite{infaas,superserve}.
For edge deployment, frameworks support model exploration and porting~\cite{retiarii,mistify}, while adaptation mechanisms use early exits~\cite{edgent} or pruning-derived model variants~\cite{nestdnn,legodnn}.
For cloud serving, systems optimize model selection, resource allocation, and request scheduling to meet latency and accuracy requirements~\cite{infaas,clockwork}.
More recently, SuperServe~\cite{superserve} extends adaptive serving to weight-sharing SuperNets by dynamically selecting SubNets in response to changing workloads.
AutoTailor complements these systems by automating SuperNet construction and enabling efficient SubNet selection, thereby is also extensible to support adaptive cloud serving.

\textbf{Optimizations for On-Device ML Inference.}
Beyond model architecture, inference efficiency can be further improved through system-level optimizations, including accelerator utilization~\cite{powernpu,heterollm,aria,vinpu}, runtime memory management~\cite{flexnn}, and kernel generation and fusion~\cite{tvm,romou,dnnfusion}. While these techniques optimize the execution of a given model on a target device, AutoTailor focuses on tailoring model architectures to device-specific characteristics. We therefore hold the inference execution stack fixed for each model–device pair, as these system-level optimizations are orthogonal to AutoTailor.

%% file: 10_conclusion.tex
\section{Conclusion}
AutoTailor makes SuperNet-based adaptive deployment practical across diverse edge devices. By automating SuperNet construction and SubNet selection with learning-free performance prediction, AutoTailor reduces the cost and complexity of architecture specialization. Across evaluated models and devices, AutoTailor reduces construction code by 10.8--34.5$\times$ and reduces predictor profiling time by 2.2--11.5$\times$, while achieving better latency–accuracy trade-offs than state-of-the-art approaches.

%% file: ref.bib
@string{conf_mobisys = {Proceedings of the ACM International Conference on Mobile Systems, Applications, and Services}}

@string{conf_mobicom = {Proceedings of the ACM International Conference on Mobile Computing and Networking}}

@string{conf_iclr = {International Conference on Learning Representations}}

@string{conf_iccv = {Proceedings of the IEEE/CVF International Conference on Computer Vision}}

@string{conf_iccvw = {Proceedings of the IEEE/CVF International Conference on Computer Vision Workshops}}

@string{conf_neurips = {Advances in Neural Information Processing Systems}}

@string{conf_icassp = {Proceedings of the IEEE International Conference on Acoustics, Speech and Signal Processing}}

@string{conf_icml = {Proceedings of the International Conference on Machine Learning}}

@string{conf_cvpr = {Proceedings of the IEEE/CVF Conference on Computer Vision and Pattern Recognition}}

@string{conf_cvpr_ieee = {Proceedings of the IEEE Conference on Computer Vision and Pattern Recognition}}

@string{conf_eccv = {Proceedings of the European Conference on Computer Vision}}

@string{conf_osdi = {Proceedings of the USENIX Symposium on Operating Systems Design and Implementation}}

@string{conf_nsdi = {Proceedings of the USENIX Symposium on Networked Systems Design and Implementation}}

@string{conf_asplos = {Proceedings of the ACM International Conference on Architectural Support for Programming Languages and Operating Systems}}

@string{conf_hpca = {Proceedings of the IEEE International Symposium on High-Performance Computer Architecture}}

@string{conf_eurosys = {Proceedings of the European Conference on Computer Systems}}

@string{conf_sensys = {Proceedings of the ACM Conference on Embedded Networked Sensor Systems}}

@string{conf_acl = {Proceedings of the Annual Meeting of the Association for Computational Linguistics}}

@string{conf_pldi = {Proceedings of the ACM SIGPLAN International Conference on Programming Language Design and Implementation}}

@string{conf_atc = {Proceedings of the USENIX Annual Technical Conference}}

@string{conf_sosp = {Proceedings of the ACM SIGOPS Symposium on Operating Systems Principles}}

@inproceedings{aria,
  author        = {Chanyoung Jung and Jeho Lee and Gunjoong Kim and Jiwon Kim and Seonghoon Park and Hojung Cha},
  title         = {{ARIA: Optimizing Vision Foundation Model Inference on Heterogeneous Mobile Processors for Augmented Reality}},
  booktitle     = conf_mobisys,
  series        = {MobiSys '25},
  pages         = {249--262},
  year          = {2025},
}

@inproceedings{panopticus,
  author        = {Jeho Lee and Chanyoung Jung and Jiwon Kim and Hojung Cha},
  title         = {{Panopticus: Omnidirectional 3D Object Detection on Resource-constrained Edge Devices}},
  booktitle     = conf_mobicom,
  series        = {MobiCom '24},
  pages         = {1207--1221},
  year          = {2024},
}

@article{edgeai2025,
  author        = {Wang, Xubin and Tang, Zhiqing and Guo, Jianxiong and Meng, Tianhui and Wang, Chenhao and Wang, Tian and Jia, Weijia},
  title         = {{Empowering Edge Intelligence: A Comprehensive Survey on On-Device AI Models}},
  journal       = {ACM Computing Surveys},
  volume        = {57},
  number        = {9},
  articleno     = {228},
  numpages      = {39},
  year          = {2025},
}

@misc{nvidiaorin,
  author        = {{NVIDIA}},
  title         = {{NVIDIA Jetson Orin}},
  year          = {2024},
  url           = {https://www.nvidia.com/en-sg/autonomous-machines/embedded-systems/jetson-orin/},
  note          = {Accessed: 2026-10-14},
}

@misc{snapdragonsoc,
  author        = {{Qualcomm}},
  title         = {{Snapdragon Overview}},
  year          = {2007},
  url           = {https://www.qualcomm.com/snapdragon/overview},
  note          = {Accessed: 2026-10-14},
}

@misc{stm32,
  author        = {{STMicroelectronics}},
  title         = {{STM32 Microcontrollers (MCUs)}},
  year          = {2007},
  url           = {https://www.st.com/en/microcontrollers-microprocessors/stm32-32-bit-arm-cortex-mcus.html},
  note          = {Accessed: 2026-10-14},
}

@inproceedings{nestdnn,
  author        = {Biyi Fang and Xiao Zeng and Mi Zhang},
  title         = {{NestDNN: Resource-Aware Multi-Tenant On-Device Deep Learning for Continuous Mobile Vision}},
  booktitle     = conf_mobicom,
  series        = {MobiCom '18},
  pages         = {115--127},
  year          = {2018},
}

@inproceedings{legodnn,
  author        = {Han, Rui and Zhang, Qinglong and Liu, Chi Harold and Wang, Guoren and Tang, Jian and Chen, Lydia Y.},
  title         = {{LegoDNN: block-grained scaling of deep neural networks for mobile vision}},
  booktitle     = conf_mobicom,
  series        = {MobiCom '21},
  pages         = {406--419},
  year          = {2021},
}

@inproceedings{adaptivenet,
  author        = {Wen, Hao and Li, Yuanchun and Zhang, Zunshuai and Jiang, Shiqi and Ye, Xiaozhou and Ouyang, Ye and Zhang, Yaqin and Liu, Yunxin},
  title         = {{AdaptiveNet: Post-deployment Neural Architecture Adaptation for Diverse Edge Environments}},
  booktitle     = conf_mobicom,
  series        = {MobiCom '23},
  pages         = {28:1--28:17},
  year          = {2023},
}

@inproceedings{ofa,
  author        = {Han Cai and Chuang Gan and Tianzhe Wang and Zhekai Zhang and Song Han},
  title         = {{Once-for-All: Train One Network and Specialize it for Efficient Deployment}},
  booktitle     = conf_iclr,
  series        = {ICLR '20},
  year          = {2020},
}

@inproceedings{elasticvit,
  author        = {Tang, Chen and Zhang, Li Lyna and Jiang, Huiqiang and Xu, Jiahang and Cao, Ting and Zhang, Quanlu and Yang, Yuqing and Wang, Zhi and Yang, Mao},
  title         = {{ElasticViT: Conflict-aware Supernet Training for Deploying Fast Vision Transformer on Diverse Mobile Devices}},
  booktitle     = conf_iccv,
  series        = {ICCV '23},
  pages         = {5829--5840},
  year          = {2023},
}

@inproceedings{dynabert,
  author        = {Lu Hou and Zhiqi Huang and Lifeng Shang and Xin Jiang and Xiao Chen and Qun Liu},
  title         = {{DynaBERT: Dynamic {BERT} with Adaptive Width and Depth}},
  booktitle     = conf_neurips,
  series        = {NeurIPS '20},
  year          = {2020},
}

@inproceedings{superdetection,
  author        = {Yuiko Sakuma and Masato Ishii and Takuya Narihira},
  title         = {{DetOFA: Efficient Training of Once-for-All Networks for Object Detection using Path Filter}},
  booktitle     = conf_iccvw,
  series        = {ICCVW '23},
  pages         = {1325--1334},
  year          = {2023},
}

@inproceedings{superasr,
  author        = {Shangguan, Yuan and Yang, Haichuan and Li, Danni and Wu, Chunyang and Fathullah, Yassir and Wang, Dilin and Dalmia, Ayushi and Krishnamoorthi, Raghuraman and Kalinli, Ozlem and Jia, Junteng and Mahadeokar, Jay and Lei, Xin and Seltzer, Mike and Chandra, Vikas},
  title         = {{TODM: Train Once Deploy Many Efficient Supernet-Based RNN-T Compression For On-Device ASR Models}},
  booktitle     = conf_icassp,
  series        = {ICASSP '24},
  pages         = {10216--10220},
  year          = {2024},
}

@inproceedings{llamaflex,
  author        = {Ruisi Cai and Saurav Muralidharan and Hongxu Yin and Zhangyang Wang and Jan Kautz and Pavlo Molchanov},
  title         = {{LLaMaFlex: Many-in-one LLMs via Generalized Pruning and Weight Sharing}},
  booktitle     = conf_iclr,
  series        = {ICLR '25},
  year          = {2025},
}

@inproceedings{compofa,
  author        = {Manas Sahni and Shreya Varshini and Alind Khare and Alexey Tumanov},
  title         = {{CompOFA: Compound Once-For-All Networks for Faster Multi-Platform Deployment}},
  booktitle     = conf_iclr,
  series        = {ICLR '21},
  year          = {2021},
}

@inproceedings{alphanet,
  author        = {Dilin Wang and Chengyue Gong and Meng Li and Qiang Liu and Vikas Chandra},
  title         = {{AlphaNet: Improved Training of Supernets with Alpha-Divergence}},
  booktitle     = conf_icml,
  series        = {ICML '21},
  pages         = {10760--10771},
  year          = {2021},
}

@inproceedings{attentivenas,
  author        = {Dilin Wang and Meng Li and Chengyue Gong and Vikas Chandra},
  title         = {{AttentiveNAS: Improving Neural Architecture Search via Attentive Sampling}},
  booktitle     = conf_cvpr,
  series        = {CVPR '21},
  pages         = {6418--6427},
  year          = {2021},
}

@inproceedings{momentumnas,
  author        = {Jeimin Jeon and Youngmin Oh and Junghyup Lee and Donghyeon Baek and Dohyung Kim and Chanho Eom and Bumsub Ham},
  title         = {{Subnet-Aware Dynamic Supernet Training for Neural Architecture Search}},
  booktitle     = conf_cvpr,
  series        = {CVPR '25},
  pages         = {30137--30146},
  year          = {2025},
}

@inproceedings{deps,
  author        = {Aditya Annavajjala and Alind Khare and Animesh Agrawal and Igor Fedorov and Hugo Latapie and Myungjin Lee and Alexey Tumanov},
  title         = {{D{\(\epsilon\)}pS: Delayed {\(\epsilon\)}-Shrinking for Faster Once-for-All Training}},
  booktitle     = conf_eccv,
  series        = {ECCV '24},
  pages         = {315--331},
  year          = {2024},
}

@inproceedings{bignas,
  author        = {Jiahui Yu and Pengchong Jin and Hanxiao Liu and Gabriel Bender and Pieter{-}Jan Kindermans and Mingxing Tan and Thomas S. Huang and Xiaodan Song and Ruoming Pang and Quoc Le},
  title         = {{BigNAS: Scaling up Neural Architecture Search with Big Single-Stage Models}},
  booktitle     = conf_eccv,
  series        = {ECCV '20},
  pages         = {702--717},
  year          = {2020},
}

@inproceedings{retiarii,
  author        = {Quanlu Zhang and Zhenhua Han and Fan Yang and Yuge Zhang and Zhe Liu and Mao Yang and Lidong Zhou},
  title         = {{Retiarii: {A} Deep Learning Exploratory-Training Framework}},
  booktitle     = conf_osdi,
  series        = {OSDI '20},
  pages         = {919--936},
  year          = {2020},
}

@inproceedings{mistify,
  author        = {Peizhen Guo and Bo Hu and Wenjun Hu},
  title         = {{Mistify: Automating {DNN} Model Porting for {On-Device} Inference at the Edge}},
  booktitle     = conf_nsdi,
  series        = {NSDI '21},
  pages         = {705--719},
  year          = {2021},
}

@inproceedings{naspipe,
  author        = {Shixiong Zhao and Fanxin Li and Xusheng Chen and Tianxiang Shen and Li Chen and Sen Wang and Nicholas Zhang and Cheng Li and Heming Cui},
  title         = {{NASPipe: High Performance and Reproducible Pipeline Parallel Supernet Training via Causal Synchronous Parallelism}},
  booktitle     = conf_asplos,
  series        = {ASPLOS '22},
  pages         = {374--387},
  year          = {2022},
}

@inproceedings{pathparallel,
  author        = {Ying Xu and Long Cheng and Xuyi Cai and Xiaohan Ma and Weiwei Chen and Lei Zhang and Ying Wang},
  title         = {{Efficient Supernet Training Using Path Parallelism}},
  booktitle     = conf_hpca,
  series        = {HPCA '23},
  pages         = {1249--1261},
  year          = {2023},
}

@inproceedings{superserve,
  author        = {Alind Khare and Dhruv Garg and Sukrit Kalra and Snigdha Grandhi and Ion Stoica and Alexey Tumanov},
  title         = {{SuperServe: Fine-Grained Inference Serving for Unpredictable Workloads}},
  booktitle     = conf_nsdi,
  series        = {NSDI '25},
  pages         = {739--758},
  year          = {2025},
}

@misc{torchnnmodule,
  author        = {{PyTorch Team}},
  title         = {{PyTorch nn.Module}},
  year          = {2025},
  url           = {https://docs.pytorch.org/docs/stable/generated/torch.nn.Module.html},
  note          = {Accessed: 2026-10-14},
}

@inproceedings{modelkeeper,
  author        = {Fan Lai and Yinwei Dai and Harsha V. Madhyastha and Mosharaf Chowdhury},
  title         = {{{ModelKeeper}: Accelerating {DNN} Training via Automated Training Warmup}},
  booktitle     = conf_nsdi,
  series        = {NSDI '23},
  pages         = {769--785},
  year          = {2023},
}

@inproceedings{optimus,
  author        = {Hong, Zicong and Lin, Jian and Guo, Song and Luo, Sifu and Chen, Wuhui and Wattenhofer, Roger and Yu, Yue},
  title         = {{Optimus: Warming Serverless ML Inference via Inter-Function Model Transformation}},
  booktitle     = conf_eurosys,
  series        = {EuroSys '24},
  pages         = {1039--1053},
  year          = {2024},
}

@inproceedings{flexnn,
  author        = {Li, Xiangyu and Li, Yuanchun and Li, Yuanzhe and Cao, Ting and Liu, Yunxin},
  title         = {{FlexNN: Efficient and Adaptive DNN Inference on Memory-Constrained Edge Devices}},
  booktitle     = conf_mobicom,
  series        = {MobiCom '24},
  pages         = {709--723},
  year          = {2024},
}

@inproceedings{romou,
  author        = {Rendong Liang and Ting Cao and Jicheng Wen and Manni Wang and Yang Wang and Jianhua Zou and Yunxin Liu},
  title         = {{Romou: rapidly generate high-performance tensor kernels for mobile GPUs}},
  booktitle     = conf_mobicom,
  series        = {MobiCom '22},
  pages         = {487--500},
  year          = {2022},
}

@inproceedings{elf,
  author        = {Zhang, Wuyang and He, Zhezhi and Liu, Luyang and Jia, Zhenhua and Liu, Yunxin and Gruteser, Marco and Raychaudhuri, Dipankar and Zhang, Yanyong},
  title         = {{Elf: accelerate high-resolution mobile deep vision with content-aware parallel offloading}},
  booktitle     = conf_mobicom,
  series        = {MobiCom '21},
  pages         = {201--214},
  numpages      = {14},
  year          = {2021},
}

@inproceedings{blastnet,
  author        = {Ling, Neiwen and Huang, Xuan and Zhao, Zhihe and Guan, Nan and Yan, Zhenyu and Xing, Guoliang},
  title         = {{BlastNet: Exploiting Duo-Blocks for Cross-Processor Real-Time DNN Inference}},
  booktitle     = conf_sensys,
  series        = {SenSys '22},
  pages         = {91--105},
  year          = {2022},
}

@inproceedings{remix,
  author        = {Jiang, Shiqi and Lin, Zhiqi and Li, Yuanchun and Shu, Yuanchao and Liu, Yunxin},
  title         = {{Flexible high-resolution object detection on edge devices with tunable latency}},
  booktitle     = conf_mobicom,
  series        = {MobiCom '21},
  pages         = {559--572},
  year          = {2021},
}

@inproceedings{ember,
  author        = {Fraternali, Francesco and Balaji, Bharathan and Sengupta, Dhiman and Hong, Dezhi and Gupta, Rajesh K.},
  title         = {{Ember: energy management of batteryless event detection sensors with deep reinforcement learning}},
  booktitle     = conf_sensys,
  series        = {SenSys '20},
  pages         = {503--516},
  year          = {2020},
}

@inproceedings{intelbeyond,
  author        = {Gobieski, Graham and Lucia, Brandon and Beckmann, Nathan},
  title         = {{Intelligence Beyond the Edge: Inference on Intermittent Embedded Systems}},
  booktitle     = conf_asplos,
  series        = {ASPLOS '19},
  pages         = {199--213},
  year          = {2019},
}

@inproceedings{llava,
  author        = {Liu, Haotian and Li, Chunyuan and Wu, Qingyang and Lee, Yong Jae},
  title         = {{Visual instruction tuning}},
  booktitle     = conf_neurips,
  series        = {NeurIPS '23},
  pages         = {34892--34916},
  year          = {2023},
}

@inproceedings{mobilenetv2,
  author        = {Mark Sandler and Andrew G. Howard and Menglong Zhu and Andrey Zhmoginov and Liang{-}Chieh Chen},
  title         = {{MobileNetV2: Inverted Residuals and Linear Bottlenecks}},
  booktitle     = conf_cvpr_ieee,
  series        = {CVPR '18},
  pages         = {4510--4520},
  year          = {2018},
}

@inproceedings{mobilenetv3,
  author        = {Andrew Howard and Ruoming Pang and Hartwig Adam and Quoc V. Le and Mark Sandler and Bo Chen and Weijun Wang and Liang{-}Chieh Chen and Mingxing Tan and Grace Chu and Vijay Vasudevan and Yukun Zhu},
  title         = {{Searching for MobileNetV3}},
  booktitle     = conf_iccv,
  series        = {ICCV '19},
  pages         = {1314--1324},
  year          = {2019},
}

@inproceedings{filterpruning,
  author        = {Hao Li and Asim Kadav and Igor Durdanovic and Hanan Samet and Hans Peter Graf},
  title         = {{Pruning Filters for Efficient ConvNets}},
  booktitle     = conf_iclr,
  series        = {ICLR '17},
  year          = {2017},
}

@inproceedings{depgraph,
  author        = {Fang, Gongfan and Ma, Xinyin and Song, Mingli and Mi, Michael Bi and Wang, Xinchao},
  title         = {{DepGraph: Towards Any Structural Pruning}},
  booktitle     = conf_cvpr,
  series        = {CVPR '23},
  pages         = {16091--16101},
  year          = {2023},
}

@inproceedings{proxylessnet,
  author        = {Han Cai and Ligeng Zhu and Song Han},
  title         = {{Proxyless{NAS}: Direct Neural Architecture Search on Target Task and Hardware}},
  booktitle     = conf_iclr,
  series        = {ICLR '19},
  year          = {2019},
}

@inproceedings{hat,
  author        = {Wang, Hanrui and Wu, Zhanghao and Liu, Zhijian and Cai, Han and Zhu, Ligeng and Gan, Chuang and Han, Song},
  title         = {{{HAT}: Hardware-Aware Transformers for Efficient Natural Language Processing}},
  booktitle     = conf_acl,
  series        = {ACL '20},
  pages         = {7675--7688},
  year          = {2020},
}

@inproceedings{autoformer,
  author        = {Chen, Minghao and Peng, Houwen and Fu, Jianlong and Ling, Haibin},
  title         = {{AutoFormer: Searching Transformers for Visual Recognition}},
  booktitle     = conf_iccv,
  series        = {ICCV '21},
  pages         = {12270--12280},
  year          = {2021},
}

@inproceedings{nnmeter,
  author        = {Li Lyna Zhang and Shihao Han and Jianyu Wei and Ningxin Zheng and Ting Cao and Yuqing Yang and Yunxin Liu},
  title         = {{nn-Meter: Towards Accurate Latency Prediction of Deep-Learning Model Inference on Diverse Edge Devices}},
  booktitle     = conf_mobisys,
  series        = {MobiSys '21},
  pages         = {81--93},
  year          = {2021},
}

@inproceedings{litepred,
  author        = {Chengquan Feng and Li Lyna Zhang and Yuanchi Liu and Jiahang Xu and Chengruidong Zhang and Zhiyuan Wang and Ting Cao and Mao Yang and Haisheng Tan},
  title         = {{{LitePred}: Transferable and Scalable Latency Prediction for {Hardware-Aware} Neural Architecture Search}},
  booktitle     = conf_nsdi,
  series        = {NSDI '24},
  pages         = {1463--1477},
  year          = {2024},
}

@inproceedings{resnet,
  author        = {He, Kaiming and Zhang, Xiangyu and Ren, Shaoqing and Sun, Jian},
  title         = {{Deep Residual Learning for Image Recognition}},
  booktitle     = conf_cvpr_ieee,
  series        = {CVPR '16},
  pages         = {770--778},
  year          = {2016},
}

@inproceedings{cnnet,
  author        = {Zhuang Liu and Hanzi Mao and Chao-Yuan Wu and Christoph Feichtenhofer and Trevor Darrell and Saining Xie},
  title         = {{A ConvNet for the 2020s}},
  booktitle     = conf_cvpr,
  series        = {CVPR '22},
  pages         = {11966--11976},
  year          = {2022},
}

@inproceedings{levit,
  author        = {Graham, Benjamin and El-Nouby, Alaaeldin and Touvron, Hugo and Stock, Pierre and Joulin, Armand and J\'egou, Herv\'e and Douze, Matthijs},
  title         = {{LeViT: A Vision Transformer in ConvNet's Clothing for Faster Inference}},
  booktitle     = conf_iccv,
  series        = {ICCV '21},
  pages         = {12259--12269},
  year          = {2021},
}

@inproceedings{vit,
  author        = {Alexey Dosovitskiy and Lucas Beyer and Alexander Kolesnikov and Dirk Weissenborn and Xiaohua Zhai and Thomas Unterthiner and Mostafa Dehghani and Matthias Minderer and Georg Heigold and Sylvain Gelly and Jakob Uszkoreit and Neil Houlsby},
  title         = {{An Image is Worth 16x16 Words: Transformers for Image Recognition at Scale}},
  booktitle     = conf_iclr,
  series        = {ICLR '21},
  year          = {2021},
}

@article{im1k,
  author        = {Olga Russakovsky and Jia Deng and Hao Su and Jonathan Krause and Sanjeev Satheesh and Sean Ma and Zhiheng Huang and Andrej Karpathy and Aditya Khosla and Michael S. Bernstein and Alexander C. Berg and Li Fei{-}Fei},
  title         = {{ImageNet Large Scale Visual Recognition Challenge}},
  journal       = {International Journal of Computer Vision},
  volume        = {115},
  number        = {3},
  pages         = {211--252},
  year          = {2015},
}

@inproceedings{tvm,
  author        = {Chen, Tianqi and Moreau, Thierry and Jiang, Ziheng and Zheng, Lianmin and Yan, Eddie and Cowan, Meghan and Shen, Haichen and Wang, Leyuan and Hu, Yuwei and Ceze, Luis and Guestrin, Carlos and Krishnamurthy, Arvind},
  title         = {{TVM: an automated end-to-end optimizing compiler for deep learning}},
  booktitle     = conf_osdi,
  series        = {OSDI '18},
  pages         = {579--594},
  year          = {2018},
}

@misc{onnx,
  author        = {{ONNX Team}},
  title         = {{Open Neural Network Exchange}},
  year          = {2021},
  url           = {https://onnx.ai/},
  note          = {Accessed: 2026-10-14},
}

@misc{gs,
  author        = {{NVIDIA Team}},
  title         = {{NVIDIA ONNX GraphSurgeon}},
  year          = {2021},
  url           = {https://docs.nvidia.com/deeplearning/tensorrt/latest/api/onnx-graphsurgeon-api.html},
  note          = {Accessed: 2026-10-14},
}

@misc{ncnn,
  author        = {{Tencent Team}},
  title         = {{Tencent NCNN}},
  year          = {2019},
  url           = {https://github.com/Tencent/ncnn/},
  note          = {Accessed: 2026-10-14},
}

@misc{pnnx,
  author        = {Nihui Zuo Zhang},
  title         = {{PNNX: PyTorch Neural Network eXchange}},
  year          = {2023},
  url           = {https://github.com/pnnx/pnnx},
  note          = {Accessed: 2026-10-14},
}

@misc{pytorch,
  author        = {{PyTorch Team}},
  title         = {{PyTorch Deep Learning Framework}},
  year          = {2016},
  url           = {https://pytorch.org/},
  note          = {Accessed: 2026-10-14},
}

@misc{timm,
  author        = {Ross Wightman},
  title         = {{PyTorch Image Models}},
  year          = {2019},
  url           = {https://github.com/rwightman/pytorch-image-models},
  note          = {Accessed: 2026-10-14},
}

@misc{onnx2torch,
  author        = {{ENOT developers} and Kalgin, Igor and Yanchenko, Arseny and Ivanov, Pyoter and Goncharenko, Alexander},
  title         = {{onnx2torch: Convert ONNX models to PyTorch}},
  year          = {2021},
  url           = {https://enot.ai/},
  note          = {Accessed: 2026-10-14},
}

@misc{nni,
  author        = {{Microsoft NNI Team}},
  title         = {{NNI: Neural Network Intelligence}},
  year          = {2022},
  url           = {https://github.com/microsoft/nni},
  note          = {Accessed: 2026-10-14},
}

@inproceedings{dnnfusion,
  author        = {Niu, Wei and Guan, Jiexiong and Wang, Yanzhi and Agrawal, Gagan and Ren, Bin},
  title         = {{DNNFusion: accelerating deep neural networks execution with advanced operator fusion}},
  booktitle     = conf_pldi,
  series        = {PLDI '21},
  pages         = {883--898},
  year          = {2021},
}

@misc{hardhat,
  author        = {{Make ML}},
  title         = {{Hard Hat Dataset}},
  year          = {2024},
  url           = {https://makeml.app/datasets/hard-hat-workers},
  note          = {Dataset mirrored at \url{https://huggingface.co/datasets/Voxel51/hard-hat-detection}; Accessed: 2026-10-14},
}

@techreport{gpt2,
  author        = {Alec Radford and Jeffrey Wu and Rewon Child and David Luan and Dario Amodei and Ilya Sutskever},
  title         = {{Language Models are Unsupervised Multitask Learners}},
  institution   = {OpenAI},
  year          = {2019},
  url           = {https://cdn.openai.com/better-language-models/language_models_are_unsupervised_multitask_learners.pdf},
}

@misc{nanogpt,
  author        = {Andrej Karpathy},
  title         = {{nanoGPT}},
  year          = {2023},
  url           = {https://github.com/karpathy/nanoGPT},
  note          = {Accessed: 2026-10-14},
}

@inproceedings{wikitext103,
  author        = {Stephen Merity and Caiming Xiong and James Bradbury and Richard Socher},
  title         = {{Pointer Sentinel Mixture Models}},
  booktitle     = conf_iclr,
  series        = {ICLR '17},
  year          = {2017},
}

@misc{yolov8,
  author        = {{Ultralytics}},
  title         = {{Ultralytics {YOLOv8}}},
  year          = {2023},
  url           = {https://github.com/ultralytics/ultralytics},
  note          = {Accessed: 2026-10-14},
}

@article{pascal-voc,
  author        = {Mark Everingham and Luc Van~Gool and Chris K. I. Williams and John Winn and Andrew Zisserman},
  title         = {{The {PASCAL Visual Object Classes (VOC)} Challenge}},
  journal       = {International Journal of Computer Vision},
  volume        = {88},
  number        = {2},
  pages         = {303--338},
  year          = {2010},
}

@article{bengio2003,
  author        = {Yoshua Bengio and R{\'e}jean Ducharme and Pascal Vincent and Christian Jauvin},
  title         = {{A Neural Probabilistic Language Model}},
  journal       = {Journal of Machine Learning Research},
  volume        = {3},
  pages         = {1137--1155},
  year          = {2003},
}

@inproceedings{greedynas,
  author        = {Shan You and Tao Huang and Mingmin Yang and Fei Wang and Chen Qian and Changshui Zhang},
  title         = {{GreedyNAS: Towards Fast One-Shot {NAS} With Greedy Supernet}},
  booktitle     = conf_cvpr,
  series        = {CVPR '20},
  pages         = {1996--2005},
  year          = {2020},
}

@inproceedings{magicat,
  author        = {Jin Xu and Xu Tan and Kaitao Song and Renqian Luo and Yichong Leng and Tao Qin and Tie{-}Yan Liu and Jian Li},
  title         = {{Analyzing and Mitigating Interference in Neural Architecture Search}},
  booktitle     = conf_icml,
  series        = {ICML '22},
  pages         = {24646--24662},
  year          = {2022},
}

@inproceedings{matformer,
  author        = {Devvrit and Kudugunta, Sneha and Kusupati, Aditya and Dettmers, Tim and Chen, Kaifeng and Dhillon, Inderjit and Tsvetkov, Yulia and Hajishirzi, Hannaneh and Kakade, Sham and Farhadi, Ali and Jain, Prateek},
  title         = {{{MatFormer}: Nested Transformer for Elastic Inference}},
  booktitle     = conf_neurips,
  series        = {NeurIPS '24},
  pages         = {140535--140564},
  year          = {2024},
}

@inproceedings{flextron,
  author        = {Ruisi Cai and Saurav Muralidharan and Greg Heinrich and Hongxu Yin and Zhangyang Wang and Jan Kautz and Pavlo Molchanov},
  title         = {{Flextron: Many-in-One Flexible Large Language Model}},
  booktitle     = conf_icml,
  series        = {ICML '24},
  pages         = {5298--5311},
  year          = {2024},
}

@misc{nemotronelastic,
  author        = {Ali Taghibakhshi and Sharath Turuvekere Sreenivas and Saurav Muralidharan and Ruisi Cai and Marcin Chochowski and Ameya Sunil Mahabaleshwarkar and Yoshi Suhara and Oluwatobi Olabiyi and Daniel Korzekwa and Mostofa Patwary and Mohammad Shoeybi and Jan Kautz and Bryan Catanzaro and Ashwath Aithal and Nima Tajbakhsh and Pavlo Molchanov},
  title         = {{{Nemotron Elastic}: Towards Efficient Many-in-One Reasoning {LLMs}}},
  year          = {2025},
  eprint        = {2511.16664},
  archiveprefix = {arXiv},
}

@misc{gemma3n,
  author        = {{Google DeepMind}},
  title         = {{Gemma 3n}},
  year          = {2025},
  url           = {https://deepmind.google/models/gemma/gemma-3n/},
  note          = {Accessed: 2026-10-14},
}

@inproceedings{infaas,
  author        = {Francisco Romero and Qian Li and Neeraja J. Yadwadkar and Christos Kozyrakis},
  title         = {{INFaaS: Automated Model-less Inference Serving}},
  booktitle     = conf_atc,
  series        = {ATC '21},
  pages         = {397--411},
  year          = {2021},
}

@inproceedings{usnet,
  author        = {Jiahui Yu and Thomas S. Huang},
  title         = {{Universally Slimmable Networks and Improved Training Techniques}},
  booktitle     = conf_iccv,
  series        = {ICCV '19},
  pages         = {1803--1811},
  year          = {2019},
}

@article{edgent,
  author        = {En Li and Liekang Zeng and Zhi Zhou and Xu Chen},
  title         = {{Edge {AI}: On-Demand Accelerating Deep Neural Network Inference via Edge Computing}},
  journal       = {IEEE Transactions on Wireless Communications},
  volume        = {19},
  number        = {1},
  pages         = {447--457},
  year          = {2020},
}

@inproceedings{clockwork,
  author        = {Arpan Gujarati and Reza Karimi and Safya Alzayat and Wei Hao and Antoine Kaufmann and Ymir Vigfusson and Jonathan Mace},
  title         = {{Serving {DNNs} like {Clockwork}: Performance Predictability from the Bottom Up}},
  booktitle     = conf_osdi,
  series        = {OSDI '20},
  pages         = {443--462},
  year          = {2020},
}

@inproceedings{powernpu,
  author        = {Daliang Xu and Hao Zhang and Liming Yang and Ruiqi Liu and Gang Huang and Mengwei Xu and Xuanzhe Liu},
  title         = {{Fast On-device {LLM} Inference with {NPUs}}},
  booktitle     = conf_asplos,
  series        = {ASPLOS '25},
  pages         = {445--462},
  year          = {2025},
}

@inproceedings{heterollm,
  author        = {Le Chen and Dahu Feng and Erhu Feng and Yingrui Wang and Rong Zhao and Yubin Xia and Pinjie Xu and Haibo Chen},
  title         = {{Characterizing Mobile {SoC} for Accelerating Heterogeneous {LLM} Inference}},
  booktitle     = conf_sosp,
  series        = {SOSP '25},
  pages         = {359--374},
  year          = {2025},
}

@inproceedings{vinpu,
  author        = {Jeho Lee and Gunjoong Kim and Chanyoung Jung and Jaehee Kim and Seonghoon Park and Hojung Cha},
  title         = {{{viNPU}: Optimizing Vision Transformer Inference on Mobile {NPUs}}},
  booktitle     = conf_eurosys,
  series        = {EuroSys '26},
  pages         = {1812--1828},
  year          = {2026},
}
